\PassOptionsToPackage{square,sort,comma,numbers}{natbib}

\documentclass{article}

\usepackage[final]{nips_2018}

\usepackage[utf8]{inputenc} 
\usepackage[T1]{fontenc}    
\usepackage{hyperref}       
\usepackage{url}            
\usepackage{booktabs}       
\usepackage{amsfonts}       
\usepackage{nicefrac}       
\usepackage{microtype}      

\usepackage{wrapfig}
\usepackage{xcolor}

\title{DropMax: Adaptive Variational Softmax}

\author{Hae Beom Lee$^{1,2}$, Juho Lee$^{3,2}$, Saehoon Kim$^2$, \bf Eunho Yang$^{1,2}$, \bf Sung Ju Hwang$^{1,2}$\\
	KAIST$^1$, AItrics$^2$, South Korea,\\ University of Oxford$^3$, United Kingdom, 
	\\\texttt{$\{$haebeom.lee, eunhoy, sjhwang82$\}$@kaist.ac.kr}\\
	\texttt{juho.lee@stats.ox.ac.uk, shkim@aitrics.com}}

\usepackage{amsmath,amsfonts}
\usepackage{amsthm}
\usepackage{amssymb,amsopn}
\usepackage{bm} 

\usepackage{graphicx}  

\usepackage{amsmath, amssymb, bm, amsthm}
\usepackage{mathrsfs} 
\usepackage{mathtools}
\usepackage{thmtools}
\usepackage{enumitem}
\usepackage{subfigure}
\usepackage{color}
\usepackage{booktabs}
\usepackage{array}
\usepackage{tabularx, booktabs, multirow}
\newcolumntype{M}[1]{>{\centering\arraybackslash}m{#1}}

\usepackage[capitalize,nameinlink]{cleveref}

\DeclareMathOperator{\Ber}{Ber}

\newcommand{\normal}{\mathcal{N}}

\newcommand{\calM}{\mathcal{M}}

\newtheorem{observation}{Observation}

\newcommand{\bw}{\mathbf{w}}
\newcommand{\bI}{\mathbf{I}}

\newcommand{\bee}{\begin{eqnarray}}
\newcommand{\eee}{\end{eqnarray}}

\newcommand{\bs}[1]{\boldsymbol{#1}} 

\newcommand{\bern}{\operatorname{Ber}}
\newcommand{\sigm}{\operatorname{sgm}}
\newcommand{\unif}{\operatorname{Unif}}
\newcommand{\kl}{\operatorname{KL}}

\newcommand{\nn}{\operatorname{NN}}

\def\loss{\mathcal{L}}

\newtheorem{cond}{Condition}

\newlength{\widebarargwidth}
\newlength{\widebarargheight}
\newlength{\widebarargdepth}
\DeclareRobustCommand{\widebar}[1]{%
	\settowidth{\widebarargwidth}{\ensuremath{#1}}%
	\settoheight{\widebarargheight}{\ensuremath{#1}}%
	\settodepth{\widebarargdepth}{\ensuremath{#1}}%
	\addtolength{\widebarargwidth}{-0.3\widebarargheight}%
	\addtolength{\widebarargwidth}{-0.3\widebarargdepth}%
	\makebox[0pt][l]{\hspace{0.15\widebarargheight}%
		\hspace{0.3\widebarargdepth}%
		\addtolength{\widebarargheight}{0.3ex}%
		\rule[\widebarargheight]{0.95\widebarargwidth}{0.1ex}}%
	{#1}}

\newcommand{\eat}[1]{}

\newcommand{\bx}{\mathbf{x}}
\newcommand{\bX}{\mathbf{X}}
\newcommand{\bY}{\mathbf{Y}}
\newcommand{\bZ}{\mathbf{Z}}
\newcommand{\bz}{\mathbf{z}}
\newcommand{\bg}{\mathbf{g}}
\newcommand{\retain}{\rho}

\newcommand{\by}{\mathbf{y}}
\newcommand{\br}{\mathbf{r}}

\newcommand{\bh}{\mathbf{h}}
\newcommand{\bo}{\mathbf{o}}
\newcommand{\bW}{\mathbf{W}}
\newcommand{\bb}{\mathbf{b}}
\newcommand{\ent}{\mathcal{H}}
\newcommand{\E}{\mathbb{E}}
\newcommand{\real}{\mathbb{R}} 

\begin{document}
	
	\maketitle
	
	\begin{abstract}
		We propose DropMax, a stochastic version of softmax classifier which at each iteration drops non-target classes according to dropout probabilities adaptively decided for each instance. Specifically, we overlay binary masking variables over class output probabilities, which are input-adaptively learned via variational inference. This stochastic regularization has an effect of building an ensemble classifier out of exponentially many classifiers with different decision boundaries. Moreover, the learning of dropout rates for non-target classes on each instance allows the classifier to focus more on classification against the most confusing classes. We validate our model on multiple public datasets for classification, on which it obtains significantly improved accuracy over the regular softmax classifier and other baselines. Further analysis of the learned dropout probabilities shows that our model indeed selects confusing classes more often when it performs classification.   
	\end{abstract}
	
	\section{Introduction}
	Deep learning models have shown impressive performances on classification tasks~\cite{alexnet,resnet,densenet}.
	However, most of the efforts thus far have been made on improving the network architecture, while the predominant choice of the final classification function remained to be the basic softmax regression.
	Relatively less research has been done here, except for few works that propose variants of softmax, such as Sampled Softmax~\cite{bengio_importance_sampling}, Spherical Softmax~\cite{spherical-softmax}, and Sparsemax~\cite{sparsemax}. However, they either do not target accuracy improvement or obtain improved accuracy only on certain limited settings.
	
	In this paper, we propose a novel variant of softmax classifier that achieves improved accuracy over the regular softmax function by leveraging the popular dropout regularization, which we refer to as \emph{DropMax}. At each stochastic gradient descent step in network training, DropMax classifier applies dropout to the exponentiations
	in the softmax function, such that we consider the true class and a random subset of other classes to learn the classifier. At each training step, this allows the classifier to be learned to solve a distinct subproblem of the given multi-class classification problem, enabling it to focus on discriminative properties of the target class relative to the sampled classes. Finally, when training is over, we can obtain an ensemble of exponentially many~\footnote{to number of classes} classifiers with different decision boundaries.
	
\begin{wrapfigure}{r}{0.5\textwidth}
		\centering
		\includegraphics[width=1.0\linewidth]{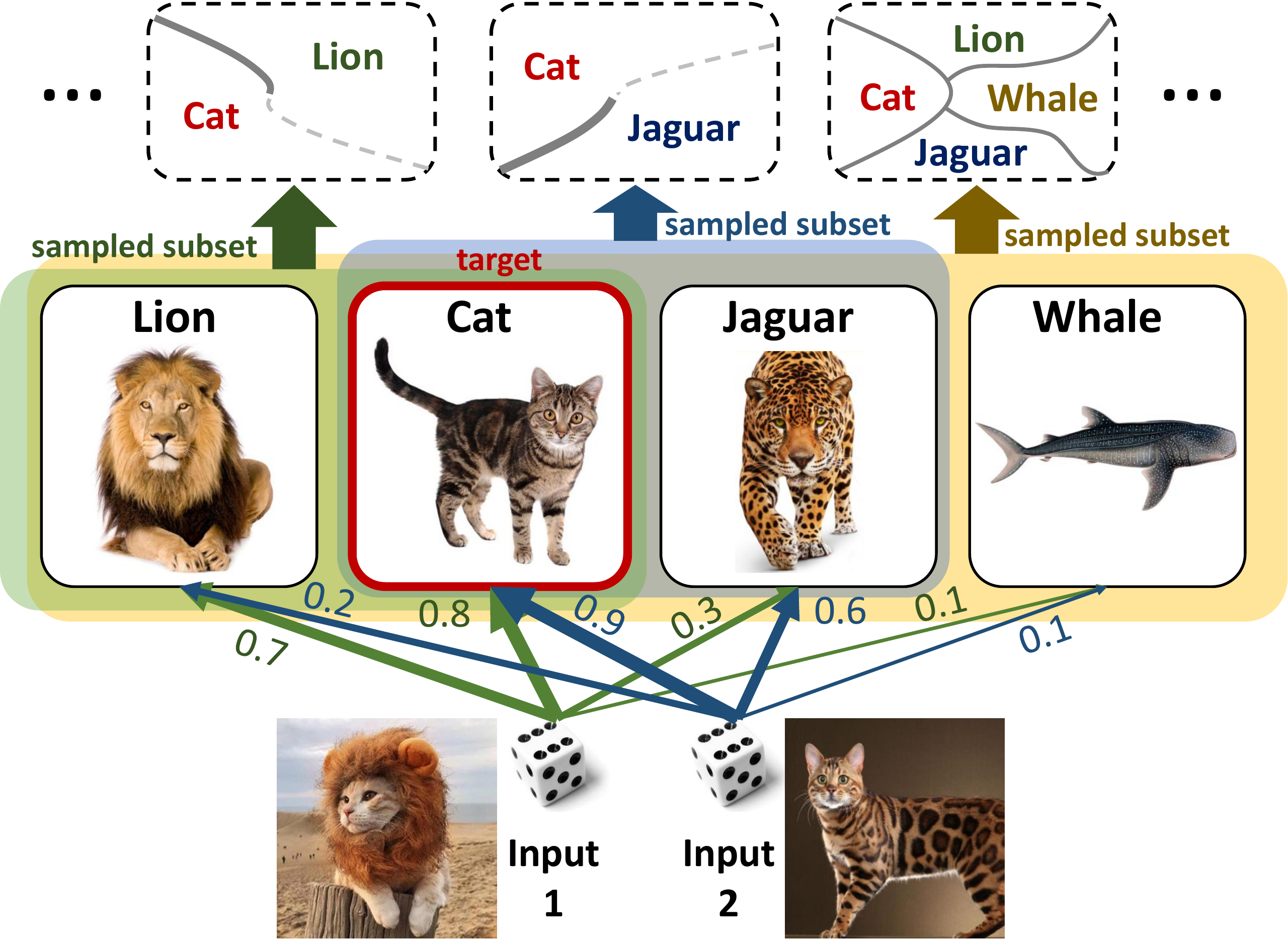}
		\caption{\textbf{Concepts.} For a given instance, classes are randomly sampled with the probabilities learned adaptively for each instance. Then the sampled subset participates in the classification.}
		\vspace{-0.1in}
		\label{concept}
	\end{wrapfigure}

	Moreover, when doing so, we further exploit the intuition that some classes could be more important than others in correct classification of each instance, as they may be confused more with the given
	instance. For example in Figure~\ref{concept},
	the instance of the class \emph{cat} on the left is likely to be more confused with class \emph{lion} because of the lion mane wig it is wearing. The \emph{cat} instance on the right, on the other hand, resembles \emph{Jaguar} due to its spots. Thus, we extend our classifier to \emph{learn} the probability of dropping non-target classes for each input instance, such that the stochastic classifier can consider
	classification against confusing classes more often than others adaptively for each input. This helps to better classify such difficult instances, which in turn will results in improving the overall classification performance. 
	
	The proposed adaptive class dropout can be also viewed as stochastic attention mechanism, that selects a subset of classes each instance should attend to in order for it to be well discriminated from any of the false classes. It also in some sense has similar effect as boosting, since learning a classifier at each iteration with randomly selected non-target classes can be seen as learning a weak classifier, which is combined into a final strong classifier that solves the complete multi-class classification problem with the weights provided by the class retain probabilities learned for each input. Our regularization is generic and can be applied even to networks on which the regular dropout is ineffective, such as ResNet, to obtain improved performance. 
	
	We validate our model on five public datasets for classification, on which it consistently obtains significant accuracy improvements over the base softmax, with noticeable improvements on fine-grained datasets ($3.38\%p$ on AWA and $7.77\%p$ on CUB dataset.)

	Our contribution is threefold:
	\begin{itemize}
		\item We propose a novel stochastic softmax function, DropMax, that randomly drops non-target classes when computing the class probability for each input instance.
		\item 
		We propose a variational inference framework to adaptively learn the dropout probability of non-target classes for each input, s.t. our stochastic classifier considers non-target classes confused with the true class of each instance more often than others.
		\item We propose a novel approach to incorporate label information into our conditional variational inference framework, 
		which yields more
		accurate posterior estimation.
	\end{itemize}
	
	\section{Related Work}
	\paragraph{Subset sampling with softmax classifier}
	Several existing work propose to consider only a partial susbset of classes to compute the softmax, as done in our work. The main motivation is on improving the efficiency of the computation,
	as matrix multiplication for computing class logits is expensive
	when there are too many classes to consider.
	For example, the number of classes (or words) often exceeds millions in language translation task.
	The common practice to tackle this challenge is to use a shortlist of $30K$ to $80K$ the most frequent target words to reduce the inherent scale of the classification problem~\cite{Bahdanau_nmt,sutskever_rare}. Further, to leverage the full vocabulary, \cite{bengio_importance_sampling} propose to calculate the importance of each word with a deterministic function and select top-$K$ among them.
	On the other hand,~\cite{sparsemax} suggest a new softmax variant that can generate sparse class probabilities, which has a similar effect to aforementioned models. Our model also works with subset of classes, but the main difference is that our model aims to improve the accuracy of the classifier, rather than improving its computational efficiency.
	
	\paragraph{Dropout variational inference} Dropout~\cite{dropout} is one of the most popular and succesful regularizers for deep neural networks. Dropout randomly drops out each neuron with a predefined probability at each iteration of a stochastic gradient descent, to achieve the effect of ensemble learning by combining exponentially many networks learned during training. Dropout can be also understood as a noise injection process~\cite{dropout_as_augmentation}, which makes the model to be robust to a small perturbation of inputs. Noise injection is also closely related to probabilistic modeling, and \cite{dropout_as_bayesian} has shown that a network trained with dropout can be seen as an approximation to deep Gaussian process. Such Bayesian understanding of dropout allows us to view model training as posterior inference, where predictive distribution is sampled by dropout at test time~\cite{what_uncertainty}. The same process can be applied to convolutional~\cite{CNN_dropout_uncertainty} and recurrent networks~\cite{RNN_dropout_uncertainty}. 
	
	\paragraph{Learning dropout probability}
	In regular dropout regularization, dropout rate is a tunable parameter that can be found via cross-validation. However, some recently proposed models allow to learn the dropout probability in the training process. Variational dropout~\cite{variational_dropout} assumes that each individual weight has independent Gaussian distribution with mean and variance, which are trained with reparameterization trick.
	Due to the central limit theorem, such Gaussian dropout is identical to the binary dropout, with much faster convergence ~\cite{dropout,fast_dropout}. \cite{sparse_variational_dropout} show that variational dropout that allows infinite variance results in sparsity, whose effect is similar to automatic relevance determination (ARD). All the aforementioned work deals with the usual posterior distribution not dependent on input at test time. On the other hand, adaptive dropout~\cite{adaptive_dropout} learns input dependent posterior at test time by overlaying binary belief network on hidden layers. Whereas approximate posterior is usually assumed to be decomposed into independent components, adaptive dropout allows us to overcome it by learning correlations between network components in the mean of input dependent posterior. Recently, \cite{concrete_dropout} proposed to train dropout probability for each layer for accurate estimation of model uncertainty, by reparameterizing Bernoulli distribution with continuous relaxation \cite{concrete_distribution}.
	
	\section{Approach}
	We first introduce the general problem setup. Suppose a dataset $\mathcal{D} = \{(\bx_i, \by_i)\}_{i=1}^N$, $\bx_i \in \real^d$, and one-hot categorical label $\by_i \in \{0,1\}^K$, with $K$ the number of classes. We will omit the index $i$ when dealing with a single datapoint. Further suppose $\bh = \nn(\bx;\omega)$, which is the last feature vector generated from an arbitrary neural network $\nn(\cdot)$ parameterized by $\omega$. Note that $\omega$ is globally optimized w.r.t. the other network components to be introduced later, and we will omit the details for brevity. We then define $K$ dimensional class logits (or scores): 
	\begin{align}
	\bo(\bx;\psi) = {\bW}^\top \bh + \bb, 
	\quad
	\psi = \{\bW, \bb\}
	\end{align}
	The original form of the softmax classifier can then be written as:
	\begin{align}
	p(\by|\bx;\psi) = \frac{\exp(o_t(\bx;\psi))}{
		\sum_k\exp(o_k(\bx;\psi))},\quad\text{where $t$ is the target class of $\bx$.} \label{eq:softmax}
	\end{align}
	
	\subsection{DropMax}
	As mentioned in the introduction, we propose to randomly drop out 
	classes
	at training phase, with the motivation of learning an ensemble of exponentially many classifiers in a single training. 
	In \eqref{eq:softmax}, one can see that class $k$ is completely excluded from the classification when $\exp(o_k)=0$, and the gradients are not back-propagated from it.
	From this observation, we randomly drop $\exp(o_1),\dots,\exp(o_K)$ based on Bernoulli trials, by introducing a dropout binary mask vector $z_k$ with \emph{retain} probability $\retain_k$, which is one minus the dropout probability for each class $k$:	
	\begin{align}
	z_k &\sim \bern(z_k;\retain_k),
	\quad p(\by|\bx,\bz;\psi) = \frac{(z_t+\varepsilon)\exp(o_t(\bf{x};\psi))}{\sum_k(z_k+\varepsilon)\exp(o_k(\bx;\psi))}
	\label{eq:pre_dropmax}
	\end{align}
	where sufficiently small $\varepsilon>0$ (e.g. $10^{-20}$) prevents the whole denominator from vanishing.
	
	However, if we drop the classes based on purely random Bernoulli trials, we may exclude the classes that are important for classification.
	Obviously, the target class $t$ of a given instance should not be dropped, but we cannot manually set the retain probabilities $\retain_{t}=1$ since the target classes differ for each instance, and more importantly, we do not know them at test time.
	We also want the retain probabilities $\retain_1,\dots,\retain_K$ to encode meaningful correlations between classes, so that the highly correlated classes may be dropped or retained together to limit the hypothesis space to a meaningful subspace.
	
	To resolve these issues, we adopt the idea of Adaptive Dropout~\cite{adaptive_dropout}, and model $\bs\retain \in [0,1]^K$ as an output of a neural network which takes 
	the last feature vector $\bh$
	as an input:
	\begin{equation}
	\bs\retain(\bx;\theta) = \sigm(\bW_\theta^\top \bh + \bb_\theta), 
	\quad
	\theta = \{\bW_\theta, \bb_\theta\}. \label{eq:retain_prob}
	\end{equation}
	By learning $\theta$, we expect these retain probabilities to be high for the target class of given inputs,
	and consider correlations between classes.  Based on this retain probability network, DropMax is defined as follows.
	\begin{equation}
	\begin{aligned}
	z_k|\bx \sim \bern(z_k;\retain_k(\bx;\theta)), \quad
	p(\by|\bx, \bz;\psi,\theta) = \frac{(z_t+\varepsilon)\exp(o_t(\bx;\psi))}
	{\sum_k (z_k+\varepsilon)\exp(o_k(\bx;\psi))}
	\label{eq:dropmax}
	\end{aligned}
	\end{equation}
	The main difference of our model from \cite{adaptive_dropout} is that,
	unlike in the adaptive dropout where the neurons of intermediate layers are dropped,
	we drop \emph{classes}. As we stated earlier, this is a critical difference, because by
	dropping classes we let the model to learn on different \emph{(sub)-problems} at each iteration, while
	in the adaptive dropout we train different \emph{models} at each iteration. Of course, our model can be extended
	to let it learn the dropout probabilities for the intermediate layers, but it is not our primary concern at this point.
	Note that DropMax can be easily applied to \emph{any} type of neural networks, such as convolutional
	neural nets or recurrent neural nets, provided that they have the softmax output for the last layer. This 
	generality is another benefit of our approach compared to the (adaptive) dropout that are reported
	to degrade the performance when used in the intermediate layers of convolutional or recurrent neural networks without careful configuration.
	
	A limitation of \cite{adaptive_dropout} is the use of heuristics to learn the dropout probabilities that may possibly result in high variance in gradients during training. To overcome this weakness, we use concrete distribution~\cite{concrete_distribution}, which is a continuous relaxation of discrete random variables that allows to back-propagate through the (relaxed) bernoulli random variables $z_k$
	to compute the gradients of $\theta$~\cite{concrete_dropout}:
		\begin{align}
	z_k = \sigm \left\{\tau^{-1} \left( \log\retain_k(\bx;\theta) - \log (1-\retain_k(\bx;\theta)) + \log u - \log (1-u)\right)\right\}\label{eq:reparam}
	\end{align}
	with $u \sim \unif (0,1)$. The temperature $\tau$ is usually set to $0.1$, which determines the degree of probability mass concentration towards $0$ and $1$.
	
	\section{Approximate Inference for DropMax}
	In this section, we describe the learning framework for DropMax. For notational simplicity, we define $\bX$, $\bY$, $\bZ$ as the concatenations of $\bx_i$, $\by_i$, and $\bz_i$ over all training instances ($i=1,\dots,N$). 
	
	\subsection{Intractable true posterior}
	We first check the form of the true posterior distribution $p(\bZ|\bX,\bY) = \prod_{i=1}^{N}p(\bz_i|\bx_i,\by_i)$. If it is tractable, then we can use exact inference algorithms such as EM to directly maximize the log-likelihood of our observation $\bY|\bX$. For each instance, the posterior distribution can be written as
	\begin{equation}
	\begin{aligned}
	p(\bz|\bx,\by) &= \frac{p(\by,\bz|\bx)}{p(\by|\bx)} = \frac{p(\by|\bx,\bz)p(\bz|\bx)}{\sum_{\bz'}p(\by|\bx,\bz')p(\bz'|\bx)}
	\label{eq:1}
	\end{aligned}
	\end{equation}
	where we let $p(\bz|\bx) = \prod_{k=1}^{K}p(z_k|\bx)$ for simplicity. However, the graphical representation of \eqref{eq:dropmax} indicates the dependencies among $z_1,\dots,z_K$ when $\by$ is observed. It means that unlike $p(\bz|\bx)$, the true posterior $p(\bz|\bx,\by)$ is not decomposable into the product of each element. Further, the denominator is the summation w.r.t. the exponentially many combinations of $\bz$, which makes the form of the true posterior even more complicated. 
	
	Thus, we suggest to use stochastic gradient variational Bayes (SGVB), which is a general framework for approximating intractable posterior of latent variables in neural network~\cite{vae, cvae}. In standard variational inference, we maximize the evidence lower bound (ELBO):
	\begin{equation}
	\begin{aligned}
	\log p&(\bY|\bX;\psi,\theta) \geq \sum_{i=1}^N \bigg\{\mathbb{E}_{q(\bz_i | \bx_i, \by_i;\phi)} \Big[ \log p(\by_i | \bz_i, \bx_i ; \psi)\Big] - \mathrm{KL}\Big[ q(\bz_i|\bx_i, \by_i;\phi) \big\Vert p(\bz_i|\bx_i;\theta)\Big]\bigg\}\label{eq:elbo}
	\end{aligned}
	\end{equation}
	where $q(\bz_i|\bx_i,\by_i;\phi)$ is our approximate posterior with a set of variational parameters $\phi$.
	 
	\subsection{Structural form of the approximate posterior}
	The probabilistic interpretation of each term in \eqref{eq:elbo} is straightforward. However, it does not tell us how to encode them into the network components. Especially, in modeling $q(\bz|\bx,\by;\phi)$, how to 
	\begin{figure}
		\vspace{-0.07in}
		\subfigure[DropMax($q=p$) (train)]{\includegraphics[height=0.15\linewidth]{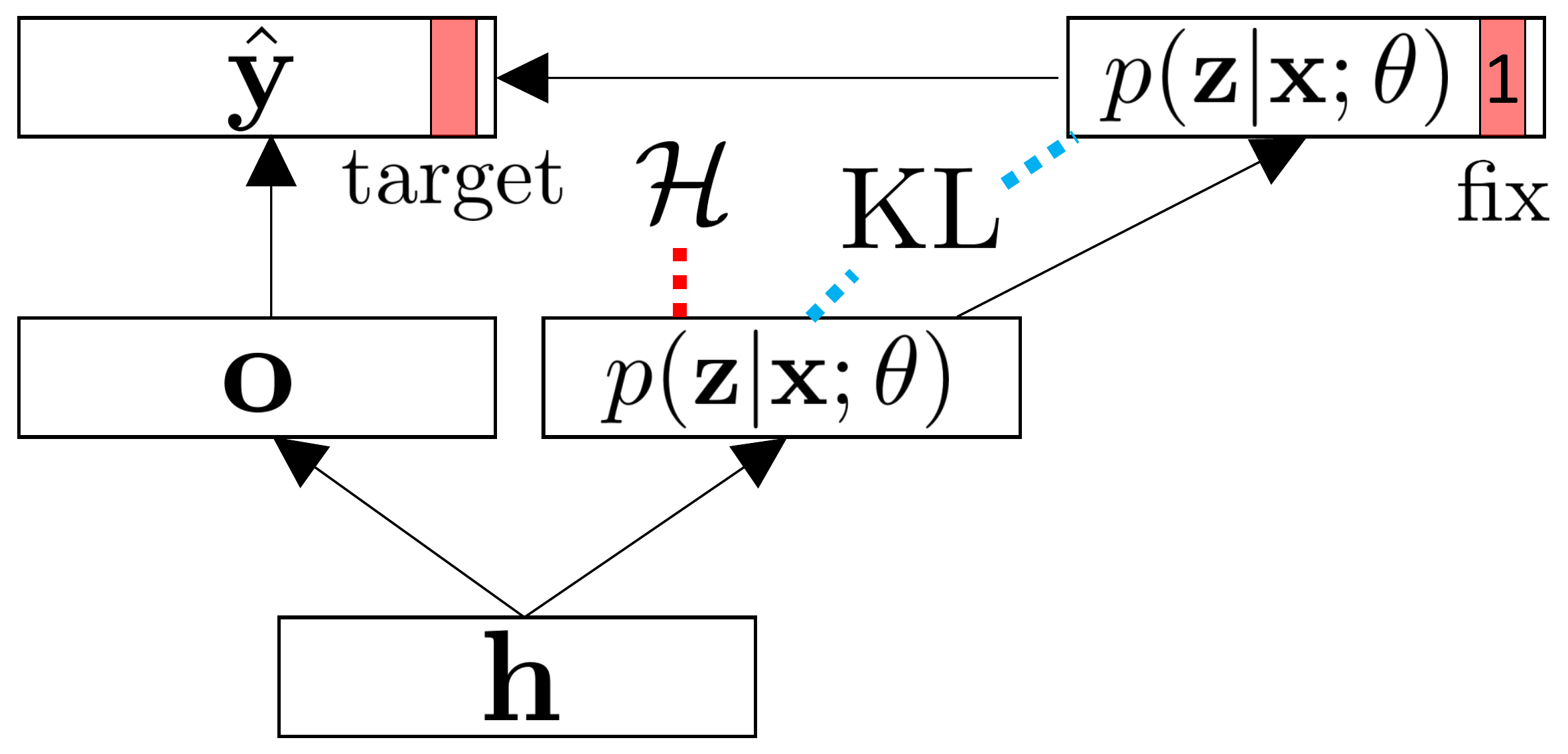}\label{qp_train_model}}
		\hfill
		\hspace{-0.25in}
		\subfigure[DropMax (train)]{\includegraphics[height=0.15\linewidth]{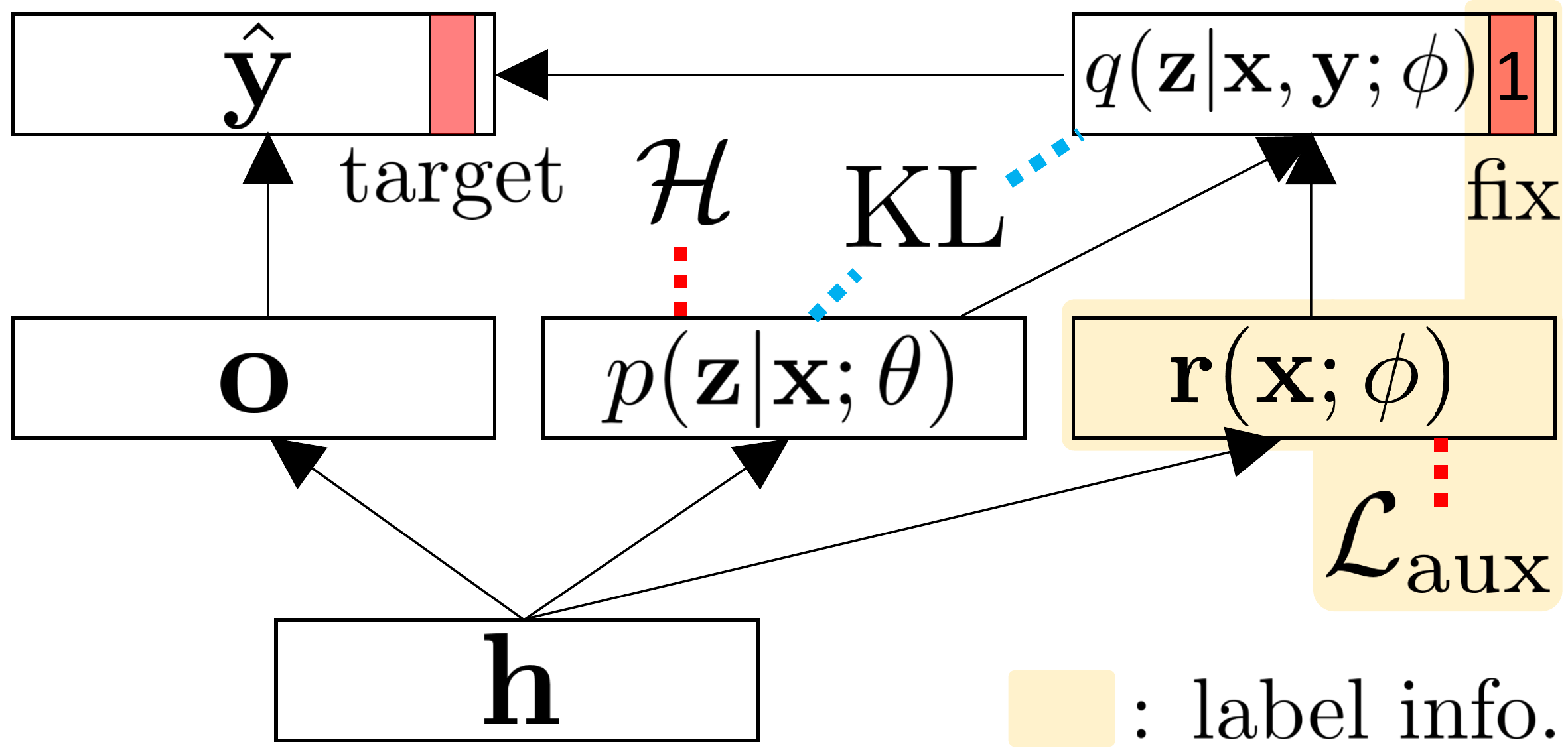}\label{train_model}}
		\hfill
		\subfigure[DropMax and DropMax($q=p$) (test)]{\includegraphics[height=0.15\linewidth]{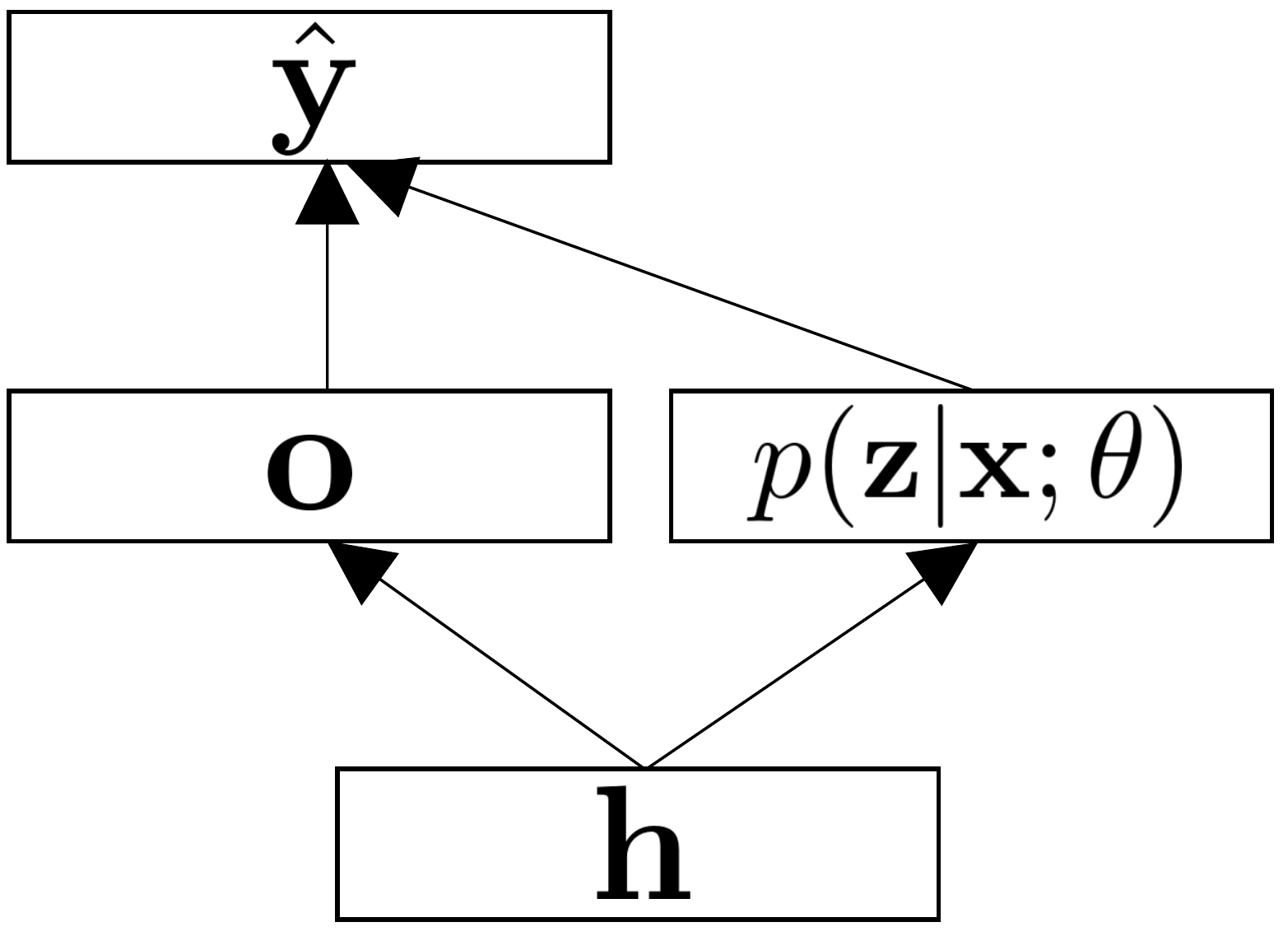}\label{test_model}}
		\vspace{-0.1in}
		\caption{\small Illustration of model architectures. (a) DropMax ($q=p$) model at training time that lets $q(\bz|\bx,\by)=p(\bz|\bx,\by;\theta)$, except that it fixes the target mask as $1$. (b) DropMax model that utilizes the label information at training time.  (c) The test-time architecture for both models.}
		\label{model}
		\vspace{-0.1in}
	\end{figure}
	utilize the label $\by$ is not a straightforward matter. \cite{cvae} suggests a simple concatenation $[\bh(\bx);\by]$ as an input to $q(\bz|\bx,\by;\phi)$, while generating a pseudo-label from $p(\bz|\bx;\theta)$ to make the pipeline of 
	training and testing network to be identical. However, it significantly increases the number of parameters of the whole network.
	On the other hand, \cite{cvae} also proposes another solution where the approximate posterior simply ignores $\by$ and share the parameters with $p(\bz|\bx;\theta)$ (Figure \ref{qp_train_model}). This method is known to be stable due to the consistency between training and testing pipeline \cite{cvae,show_attend_tell,what_uncertainty}. However, we empirically found that this approach produces suboptimal results for DropMax since it yields inaccurate approximated posterior.
	
	Our novel approach to solve the problem starts from the observation that the relationship between $\bz$ and $\by$ is relatively simple in DropMax \eqref{eq:dropmax}, unlike the case where latent variables are assumed at lower layers. In this case, even though a closed form of true posterior is not available, we can capture a few important property of it and encode them into the approximate posterior.
	
	The first step is to encode the structural form of the true posterior \eqref{eq:1}, which is decomposable into two factors: 1) the factor dependent only on $\bx$, and 2) the factor dependent on both $\bx$ and $\by$.
	\begin{align}
	p(\bz|\bx,\by) = \underbrace{p(\bz|\bx)}_\mathcal{A}\times \underbrace{p(\by|\bz,\bx)/p(\by|\bx)}_\mathcal{B}.	\label{eq:fact}
	\end{align}
	The key idea is that the factor $\mathcal{B}$ can be interpreted as the rescaling factor from the unlabeled posterior $p(\bz|\bx)$, which takes $\bx$ and $\by$ as inputs. In doing so, we model the approximate posterior $q(\bz|\bx,\by)$ with two pipelines. Each of them corresponds to: $\mathcal{A}$ without label, which we already have defined as $p(\bz|\bx;\theta) = \bern(\bz;\sigm(\bW_\theta^\top\bh + \bb_\theta))$ in \eqref{eq:retain_prob}, and $\mathcal{B}$ with label, which we discuss below.
	
	$\mathcal{B}$ is able to scale up or down $\mathcal{A}$, but at the same time should bound the resultant posterior $p(\bz|\bx,\by)$ in the range of $[0,1]^K$. To model $\mathcal{B}$ with network components, we simply add to the logit of $\mathcal{A}$, a vector $\br \in \real^K$ taking $\bx$ as an input (we will defer the description on how to use $\by$ to the next subsection). Then we squash it again in the range of $[0,1]$ (Note that addition in the logit level is multiplicative): 
	\begin{equation}
	\begin{aligned}
	\bg(\bx;\phi) = \sigm(\widebar\bW_\theta^\top \bh + \widebar\bb_\theta + 
	\br(\bx;\phi)
	),\quad \br(\bx;\phi) = \bW_\phi^\top \bh + \bb_\phi \label{eq:q}.
	\end{aligned}
	\end{equation}
	where $\bg(\bx;\phi)$ is the main ingredient of our approximate posterior in \eqref{eq:approx_post}, and $\phi=\{ \bW_\phi,\bb_\phi \}$ is variational parameter. 
	$\widebar\bW_\theta$ and $\widebar\bb_\theta$ denote that stop-gradients are applied to them, to make sure $\bg(\bx;\phi)$ is only parameterized by the variational parameter $\phi$, which is important for properly definiting the variational distribution.
	Next we discuss how to encode $\by$ into $\br(\bx;\phi)$ and $\bg(\bx;\phi)$, to finialize the approximate posterior $q(\bz|\bx,\by;\phi)$. 
	
	\subsection{Encoding the label information}
	Our modeling choice for encoding $\by$ is based on the following observations.
	\begin{observation}
		If we are given $(\bx, \by)$ and consider $z_1,\dots,z_K$ one by one, then $z_t$ is positively correlated with the DropMax likelihood $p(\by|\bx,\bz)$ in \eqref{eq:dropmax}, while $z_{k \neq t}$ is negatively correlated with it.
	\end{observation}
	\begin{observation}
		The true posterior of the target retain probability $p(z_t=1|\bx,\by)$ is $1$, if we exclude the case $z_1=z_2=\dots=z_K=0$, i.e. the retain probability for every class is $0$.
	\end{observation}
	One can easily verify the observation 1: the likelihood will increase if we attend the target more, and vice versa.
	We encode this observation as follows. Noting that the likelihood $p(\by|\bx,\bz)$ is in general maximized over the training instances, the factor $\mathcal{B}$ in \eqref{eq:fact} involves $p(\by|\bx,\bz)$ and should behave consistently (as in observation 1). Toward this, each $r_t(\bx;\phi)$ and $r_{k \neq t}(\bx;\phi)$ should be maximized and minimized respectively. 
	We achieve this by minimizing the cross-entropy for $\sigm(\br(\bx;\phi))$ across the training instances:
	\begin{equation}
	\begin{aligned}
	\loss_{\text{aux}}(\phi) = -\sum_{i=1}^N\sum_{k=1}^K \Big\{ y_{i,k}\log \sigm(r_k(\bx_i;\phi)) + (1-y_{i,k})\log(1-\sigm(r_k(\bx_i;\phi))) \Big\}
	\label{eq:rloss}
	\end{aligned}
	\end{equation}
	The observation 2 says that $\bz_{\backslash t} \neq \mathbf{0} \rightarrow z_t = 1$  given $\by$. Thus, simply ignoring the case $\bz_{\backslash t} = \mathbf{0}$ and fixing $q(z_t|\bx,\by;\phi)=\bern(z_t;1)$ is a close approximation of $p(z_t|\bx,\by)$, especially under mean-field assumption (see the Appendix A for justification).
	Hence, our final approximate posterior is given as:
	\begin{equation}
	\begin{aligned}
	q(\bz|\bx,\by;\phi) = \bern(z_t;1)\prod_{k\neq t}\bern\left(z_k;g_k(\bx;\phi)\right)\label{eq:approx_post}.
	\end{aligned}
	\end{equation}
	See Figure \ref{train_model} and \ref{test_model} for the illustration of the model architecture.
	
	\subsection{Regularized variational inference}\label{section:regularized_variational_inference}
	One last critical issue in optimizing the ELBO \eqref{eq:elbo} is that $p(\bz|\bx;\theta)$ collapses into $q(\bz|\bx,\by;\phi)$ too easily, as $p(\bz|\bx;\theta)$ is parameteric with input $\bx$. Preventing it is crucial for $\bz$ to generalize well on a test instance $\bx^*$, because $\bz$ is sampled from $p(\bz|\bx^*;\theta)$ at test time (Figure \ref{test_model}).
	We empirically found that imposing some prior (e.g. zero-mean gaussian or laplace prior) to $\theta=\{\bW_\theta,\bb_\theta\}$ was not effective in preventing this behavior. 
	(The situation is different from VAE \cite{vae} where the prior of latent code $p(\bz)$ is commonly set to gaussian with no trainable parameters (i.e. $\normal(\mathbf{0},\lambda\bI)$).)
	
	We propose to remove weight decay for $\theta$ and apply an entropy regularizer directly to $p(\bz|\bx;\theta)$. We empirically found that the method works well without any scaling hyperparameters. 
	\begin{align}
	\ent(p(\bz|\bx;\theta)) = \sum_k \rho_k \log \rho_k + (1-\rho_k) \log (1-\rho_k)\label{eq:entropy}
	\end{align}
	We are now equipped with all the essential components. The KL divergence and the final minimization objective are given as:
	\begin{equation}
	\begin{aligned}
	&\kl[q(\bz|\bx,\by;\phi)||p(\bz|\bx;\theta)] = \sum_k \Bigg\{\mathbb{I}_{\{k=t\}}\log \frac{1}{\rho_k} + \mathbb{I}_{\{k\neq t\}}\bigg(g_k\log\frac{g_k}{\rho_k} + (1-g_k) \log\frac{1-g_k}{1-\rho_k} \bigg) \Bigg\}
	\nonumber
	\\
	&\loss(\psi,\theta,\phi) = \sum_{i=1}^N \bigg[-\frac{1}{S} \sum_{s=1}^{S} \log p(y_i|\bx_i,\bz_i^{(s)};\psi)
	+ \kl[q(\bz_i|\bx_i,\by_i;\phi)||p(\bz_i|\bx_i;\theta)] - \ent \bigg] + \loss_{\text{aux}}
	\nonumber
	\end{aligned}
	\end{equation}
	where $\bz_i^{(s)} \sim q(\bz_i|\bx_i,\by_i;\phi)$ and $S=1$ as usual. 
	Figure \ref{model}(b) and (c) illustrate the model architectures for training and testing respectively.
	
	When testing, 
	we can perform Monte-Carlo sampling:
	\begin{align}
	p(\by^*|
	\bx^*)=\E_{\bz}[p(\by^*|\bx^*,\bz)] \approx \frac{1}{S}\sum_{s=1}^S p(\by^*|\bx^*,\bz^{(s)}),\quad \bz^{(s)} \sim p(\bz|\bx^*;\theta).\label{eq:test_ensemble}
	\end{align}
	Alternatively, we can approximate the expectation as
	\begin{align}
	p(\by^*|
	\bx^*)=\E_{\bz}[p(\by^*|\bx^*,\bz)]\approx p(\by^*|\bx^*,\E[\bz|\bx^*])=p(\by^*|\bx^*,\rho(\bx^*;\theta)),\label{eq:test_approx}
	\end{align}
	which is a common practice for many practitioners. We report test error based on \eqref{eq:test_approx}.

	\section{Experiments}\label{experiments}
	\paragraph{Baselines and our models} We first introduce relevant baselines and our models.
	
	\textbf{1) Base Softmax.} The baseline CNN network with softmax, that only uses the hidden unit dropout at fully connected layers, or no dropout regularization at all.
	
	\textbf{2) Sparsemax.} Base network with Sparsemax loss proposed by \cite{sparsemax}, which produces sparse class probabilities.
	
	\textbf{3) Sampled Softmax.} Base network with sampled softmax \cite{bengio_importance_sampling}. Sampling function $Q(\by|\bx)$ is uniformly distributed during training. We tune the number of sampled classes among
	$\{20\%,40\%,60\%\}$ of total classes, while the target class is always selected. Test is done with \eqref{eq:softmax}.
	
	\textbf{4) Random Dropmax.} A baseline that randomly drops out non-target classes with a predefined retain probability $\rho \in \{0.2, 0.4, 0.6\}$ at training time. For learning stability, the target class is not dropped out during training. Test is done with the softmax function \eqref{eq:softmax}, without sampling the dropout masks.
	
	\textbf{5) Deterministic Attention.}
	Softmax with deterministic sigmoid attentions multiplied at the exponentiations. The attention probabilities are generated from the last feature vector $\bh$ with additional weights and biases, similarly to \eqref{eq:retain_prob}.

	
	\textbf{6) Deterministic DropMax.} This model is the same with Deterministic Attention, except that it is trained in a supervised manner with true class labels. With such consideration of labels when training the attention generating network, the model can be viewed as a deterministic version of DropMax. 
	
	\textbf{7) DropMax ($q=p$).} A variant of DropMax where we let $q(\bz|\bx,\by)=p(\bz|\bx;\theta)$ except that we fix $q(z_t|\bx,\by)=\bern(z_t;1)$ as in \eqref{eq:approx_post} for learning stability. The corresponding $\kl[q\|p]$ can be easily computed. The auxiliary loss term $\loss_\text{aux}$ is removed and the entropy term $\ent$ is scaled with a hyperparameter $\gamma \in \{1, 0.1, 0.01\}$ (See Figure \ref{qp_train_model}).
	
	\textbf{8) DropMax.} Our adaptive stochastic softmax, where each class is dropped out with input dependent probabilities trained from the data. No hyperparameters are needed for scaling each term.
	
	We implemented DropMax using Tensorflow~\cite{tensorflow} framework. The source codes are available at \url{https://github.com/haebeom-lee/dropmax}.
	
		\begin{table}[t]
		\small
		\caption{\textbf{Classification performance in test error (\%).} 
			The reported numbers are mean and standard errors with $95\%$ confidence interval over 5 runs.
		}
		\vspace{-0.1in}
		\begin{center}
			\resizebox{\textwidth}{!}{
				\begin{tabular}{cccccccc}
					Models & M-$1K$ & M-$5K$ & M-$55K$ & C10 & C100 & AWA & CUB \\
					\hline
					\hline
					Base Softmax & $7.09_{\pm 0.46}$ & $2.13_{\pm0.21}$ & $0.65_{\pm0.04}$ & $7.90_{\pm0.21}$ & $30.60_{\pm0.12}$ & $30.29_{\pm 0.80}$ & $48.84_{\pm 0.85}$ \\
					Sparsemax~\citep{sparsemax}	& $6.57_{\pm0.17}$ & $2.05_{\pm0.18}$ & $0.75_{\pm0.06}$ & $7.90_{\pm0.28}$ & $31.41_{\pm0.16}$ & $36.06_{\pm 0.64}$ & $64.41_{\pm 1.12}$ \\
					Sampled Softmax~\citep{bengio_importance_sampling} 	& $7.36_{\pm0.22}$ & $2.31_{\pm0.14}$ & $0.66_{\pm0.04}$ & $7.98_{\pm0.24}$ & $30.87_{\pm0.19}$ & $29.81_{\pm 0.45}$ & $49.90_{\pm 0.56}$ \\
					Random Dropmax & $7.19_{\pm0.57}$ & $2.23_{\pm0.19}$ & $0.68_{\pm0.07}$ & $8.21_{\pm0.08}$ & $30.78_{\pm0.28}$ & $31.11_{\pm 0.54}$ & $48.87_{\pm 0.79}$ \\
					Deterministic Attention	& $6.91_{\pm0.46}$ & $2.03_{\pm0.11}$ & $0.69_{\pm0.05}$ & $7.87_{\pm0.24}$ & $30.60_{\pm0.21}$ & $30.98_{\pm 0.66}$ & $49.97_{\pm 0.32}$ \\
					\hline	
					Deterministic DropMax & $6.30_{\pm0.64}$ & $1.89_{\pm0.04}$ & $0.64_{\pm0.05}$ &  $7.84_{\pm0.14}$ & $30.55_{\pm0.51}$ & $\bf26.22_{\pm0.76}$ & $47.35_{\pm 0.42}$ \\
					DropMax ($q=p$) & $7.52_{\pm 0.26}$ & $2.05_{\pm 0.07}$ & $0.63_{\pm 0.02}$ & $7.80_{\pm 0.22}$ & $29.98_{\pm 0.35}$ & $29.27_{\pm 1.19}$ & $42.08_{\pm 0.94}$ \\
					DropMax & $\bf5.32_{\pm0.09}$ & $\bf1.64_{\pm0.08}$ & $\bf0.59_{\pm0.04}$ & $\bf7.67_{\pm0.11}$ & $\bf29.87_{\pm0.36}$ & $26.91_{\pm 0.54}$ & $\bf41.07_{\pm 0.57}$
				\end{tabular}\label{maintable}
				}
		\end{center}
		\vspace{-0.15in}
	\end{table}
	
	\paragraph{Datasets and base networks} We validate our method on multiple public datasets for classification, with different network architecture for each dataset.
	
	\textbf{1) MNIST.} This dataset~\cite{mnist} consists of $60,000$ images that describe hand-written digits from $0$ to $9$. 
	We experiment with varying number of training instances: $1K$, $5K$, and $55K$. The validation and test set has $5K$ and $10K$ instances, respectively. 
	As for the base network, we use the CNN provided in the Tensorflow Tutorial,
	which has a similar structure to LeNet. 
	
	\textbf{2) CIFAR-10.} This dataset~\cite{cifar} consists of $10$ generic object classes, which for each class has $5000$ images for training and $1000$ images for test. We use ResNet-34~\cite{resnet} as the base network.
	
	\textbf{3) CIFAR-100.} This dataset consists of $100$ object classes. It has $500$ images for training and $100$ images are for test for each class. We use ResNet-34 as the base network.
	
	\textbf{4) AWA.} This is a dataset for classifying different animal species~\cite{lampert-attributes}, that contains $30,475$ images from $50$ animal classes such as~\emph{cow}, \emph{fox}, and \emph{humpback whale}. For each class, we used $50$ images for test, while rest of the images are used as training set. 
	We use ResNet-18 as the base network.
	
	\textbf{5) CUB-200-2011.} This dataset~\cite{cub} consists of $200$ bird classes such as~\emph{Black footed albatross}, \emph{Rusty blackbird}, and \emph{Eastern towhee}. It has $5994$ training images and $5794$ test images, which is quite small compared to the number of classes.  
	We only use the class label for the classification. We use ResNet-18 as the base network.
	
	As AWA and CUB datasets are subsets of ImageNet-1K dataset, for those datasets we do not use a pretained model but train from scratch.
	The experimental setup is available in the Appendix C.
	
	\subsection{Quantitative Evaluation}
	\paragraph{Multi-class classification.}
	We report the classification performances of our models and the baselines in Table~\ref{maintable}.
	The results show that the variants of softmax function such as Sparsemax and Sampled Softmax perform similarly to the original softmax function (or worse). Class Dropout also performs worse due to the inconsistency between train and test time handling of the dropout probabilities for target class. Deterministic Attention also performs similarly to all the previous baselines. Interestingly, Deterministic DropMax with supervised learning of attention mechanism improves the performance over the base soft classifier, which suggests that such combination of a multi-class and multi-label classifier could be somewhat beneficial. However, the improvements are marginal except on the AWA dataset, because the gating function also lacks proper regularization and thus yields very sharp attention probabilities. DropMax $(q=p)$ has an entropy regularizer to address this issue, but the model obtains suboptimal performance due to the inaccurate posterior estimation.
	
	On the other hand, the gating function of DropMax is optimally regularized to make a \emph{crude} selection of candidate classes via the proposed variational inference framework, and shows consistent and significant improvements over the baselines across all datasets. DropMax also obtains noticeably higher accuracy gains on AWA and CUB dataset that are fine-grained, with $3.38\%p$ and $7.77\%p$ improvements, as these fine-grained datasets contain many ambiguous instances that can be effectively handled by DropMax with its focus on the most confusing classes. On MNIST dataset, we also observe that the DropMax is more effective when the number of training instances is small. We attribute this to the effect of stochastic regularization that effectively prevents overfitting. This is also a general advantage of Bayesian learning as well.
	
	
	\paragraph{Convergence rate.}
	\begin{wrapfigure}{r}{0.55\textwidth}
		\vspace{-0.23in}
		\centering
		\hfill
		\subfigure[MNIST-55K]{\includegraphics[width=0.49\linewidth]{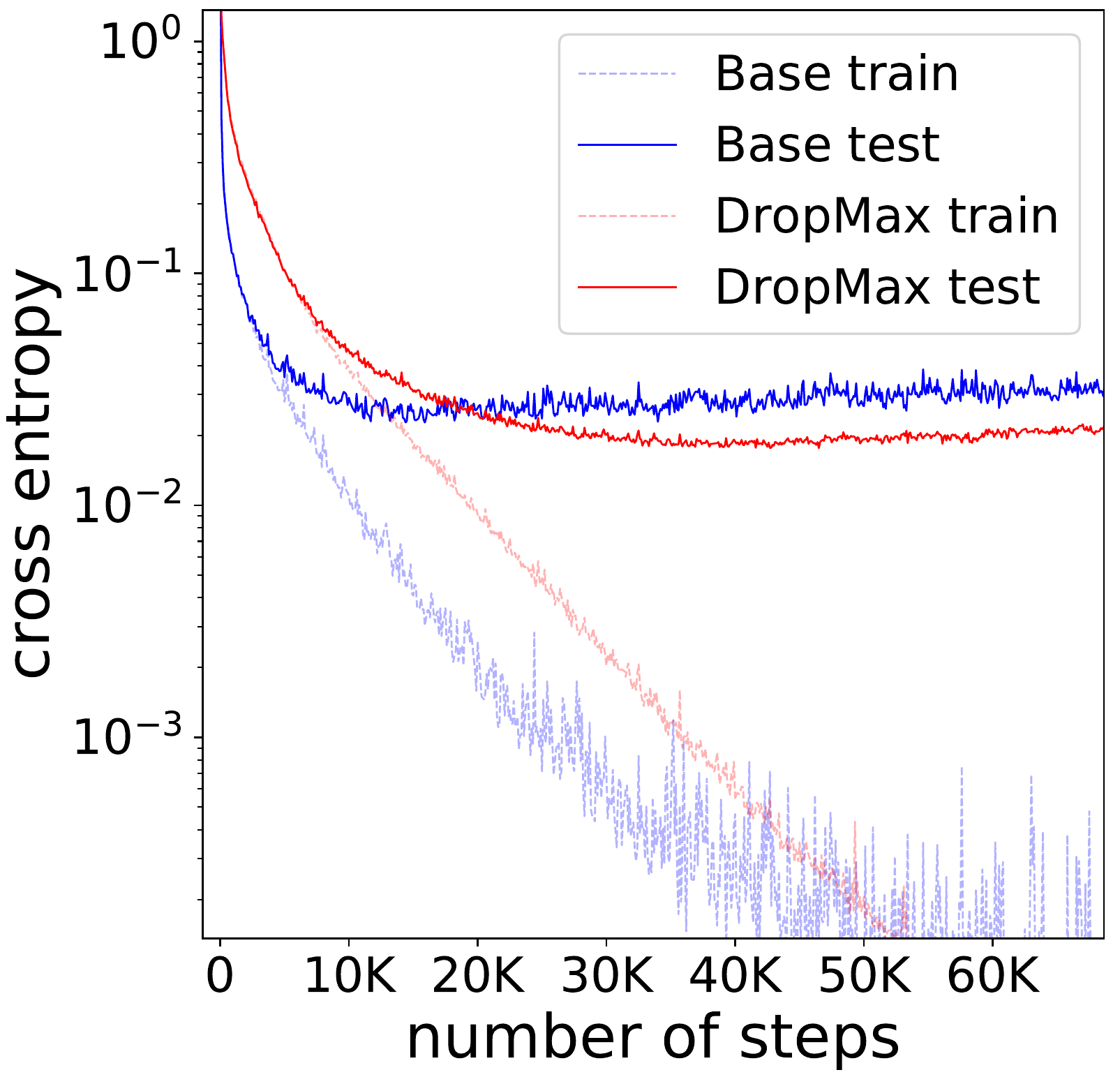}\label{curve1}}
		\hfill
		\subfigure[CIFAR-100]{\includegraphics[width=0.49\linewidth]{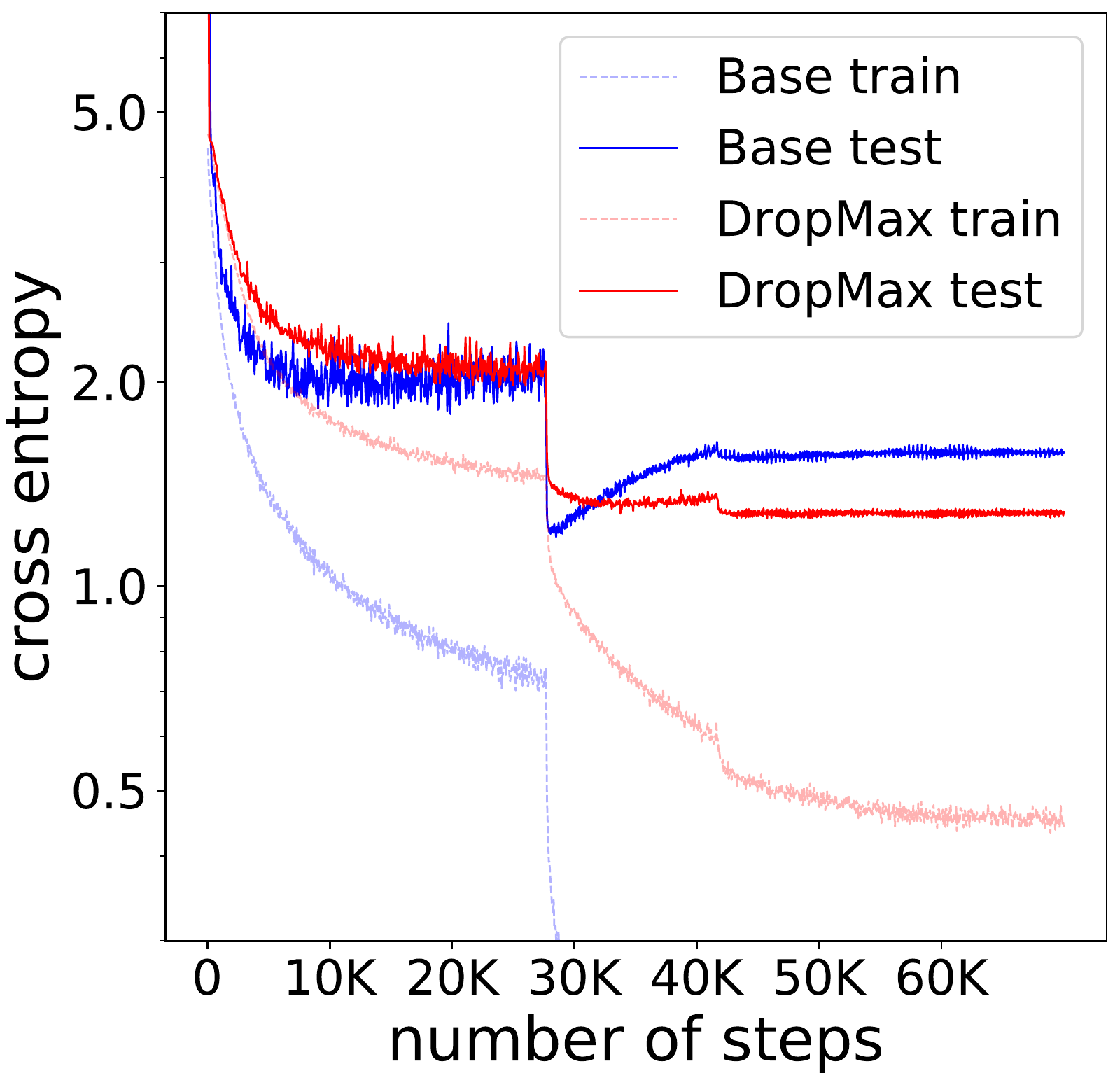}\label{curve2}}
		\hfill
		\vspace{-0.19in}
		\caption{\small Convergence plots}
		\vspace{-0.2in}
		\label{convergence}
	\end{wrapfigure}
	We examine the convergence rate of our model against the base network with regular softmax function. Figure~\ref{convergence} shows the plots of cross entropy loss computed at each training step on MNIST-55K and CIFAR-100. To reduce the variance of $\bz$, we plot with $\bs\rho$ instead (training is done with $\bz$).
	DropMax shows slightly lower convergence rate, but the test loss is significantly improved, effectively preventing overfitting. Moreover, the learning curve of DropMax is more stable than that of regular softmax (see Appendix B for more discussion). 
	
	\subsection{Qualitative Analysis}
	We further perform qualitative analysis of our model to see how exactly it works and where the accuracy improvements come from.
	
	\begin{figure*}[t]
		\hfill
		\subfigure[$\bs\rho$ (easy)]{\includegraphics[height=3.65cm]{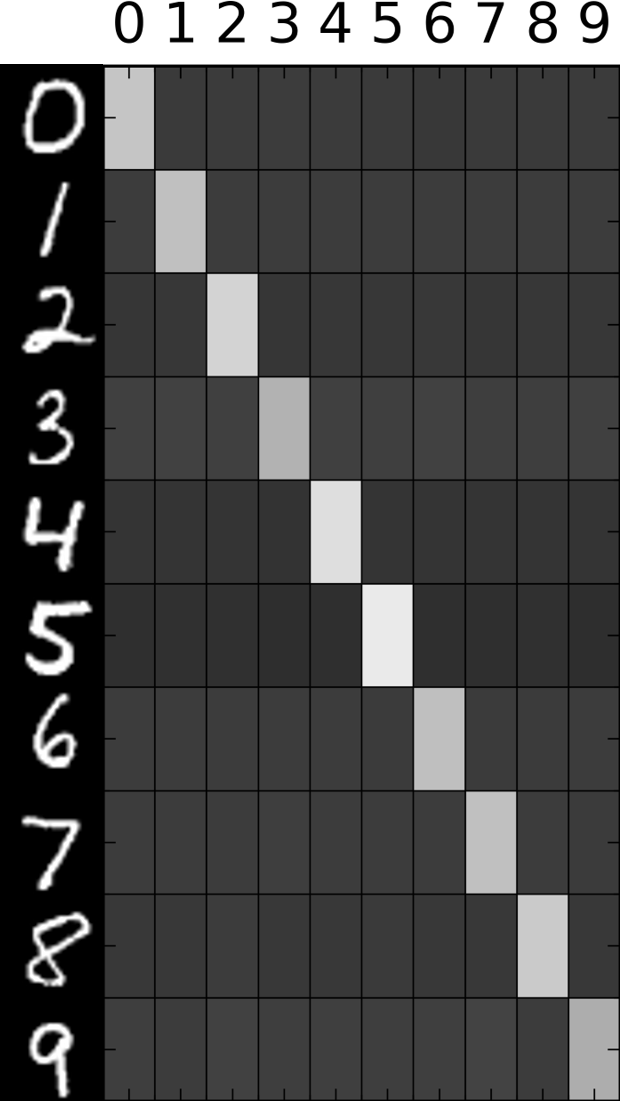}\label{easy_examples}}
		\hfill
		\subfigure[$\bs\rho$ (hard)]{\includegraphics[height=3.65cm]{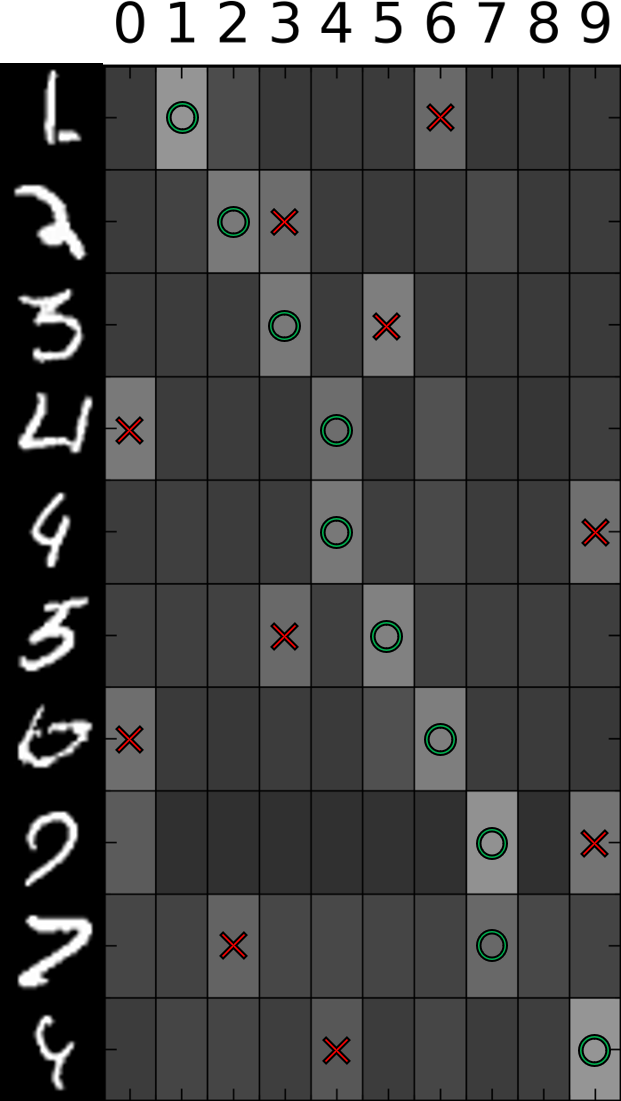}\label{p_examples}} 
		\hfill
		\subfigure[$q$ (hard)]{\includegraphics[height=3.65cm]{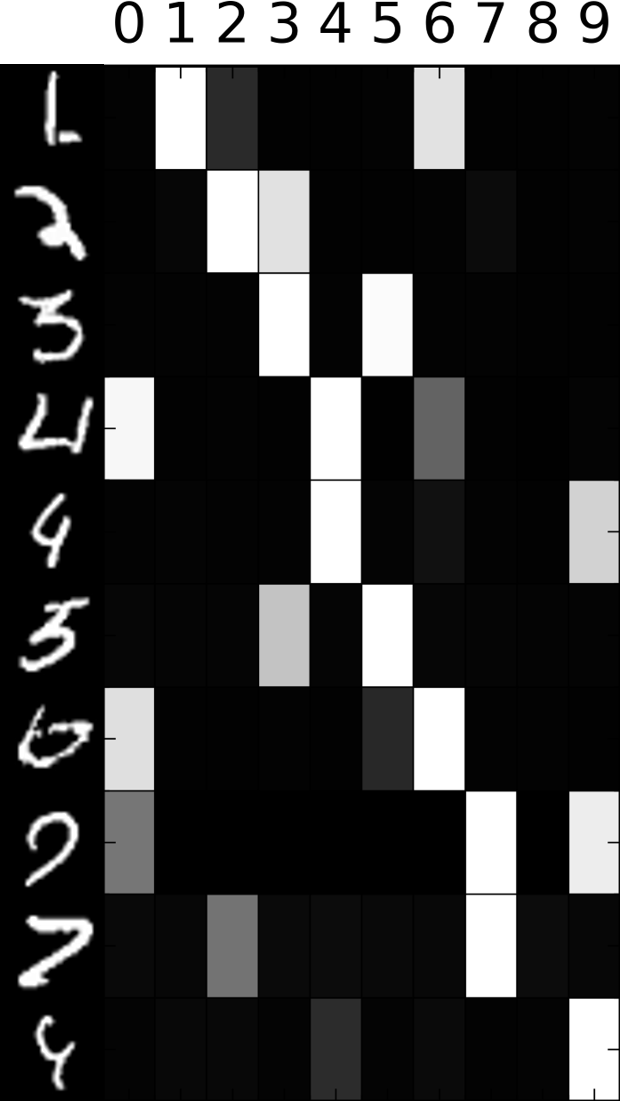}\label{q_examples}} 
		\hfill
		\subfigure[$\bz^{(s)}$ (hard)]{\includegraphics[height=3.65cm]{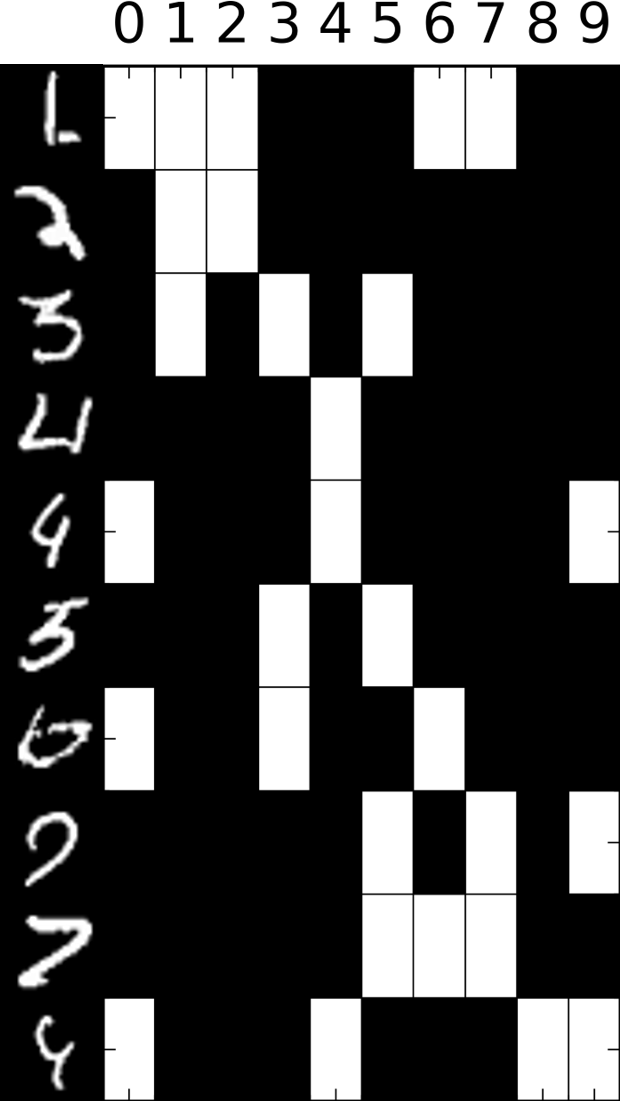}\label{z_examples}}  
		\subfigure[$\bar{\bs\rho}$ (hard)]{\includegraphics[height=3.65cm]{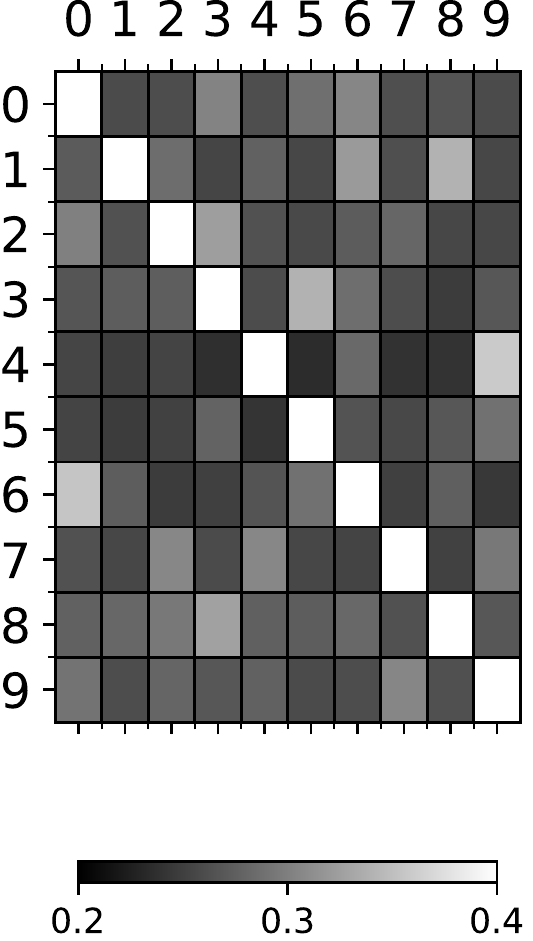}\label{B}}
		\vspace{-0.1in}
		\subfigure[$p(\by|\bx)$]{\includegraphics[height=3.6cm]{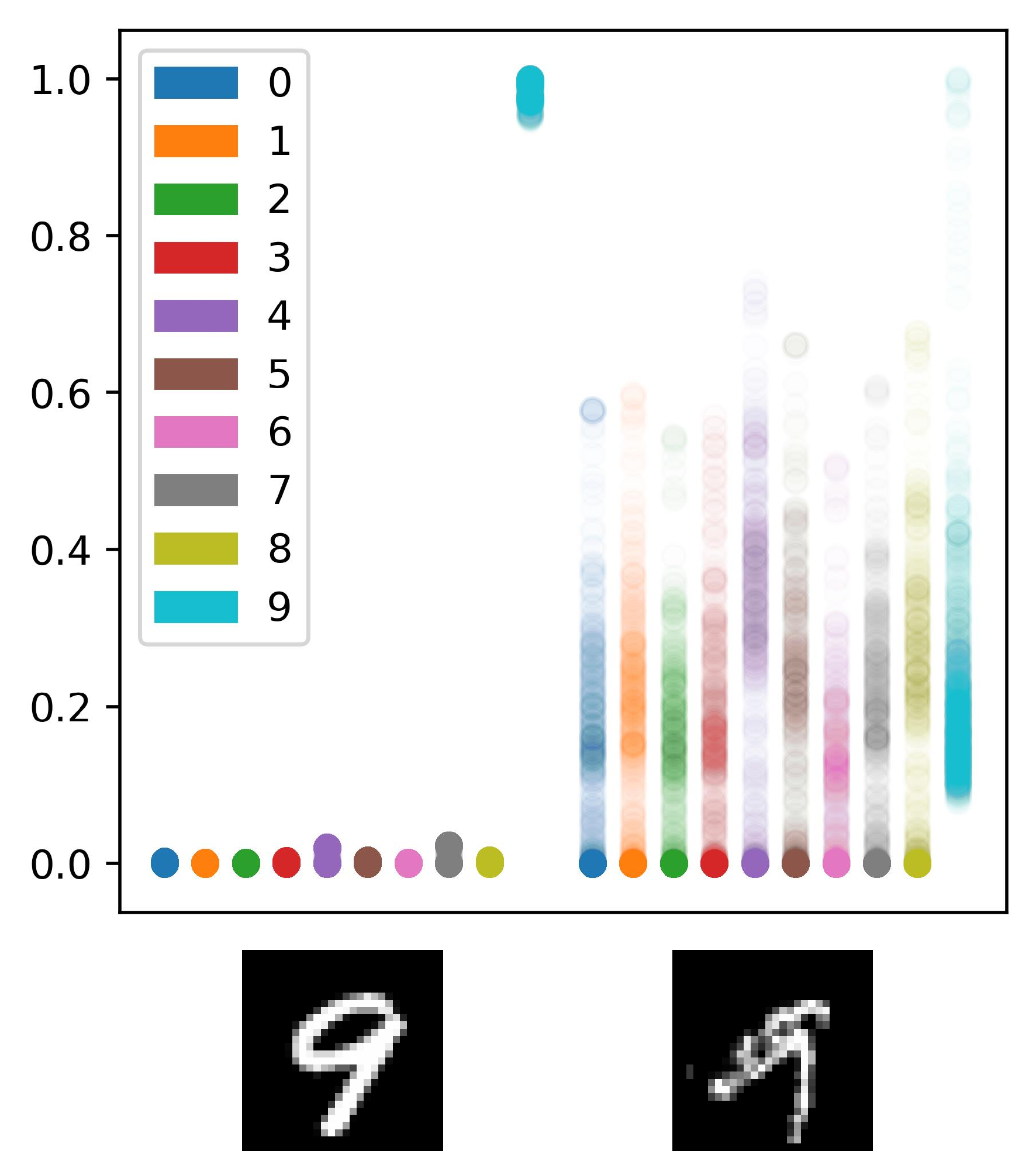}\label{uncertainty}}
		\caption{\small Visualization of class dropout probabilities for example test instances from MNIST-1K dataset. (a) and (b) shows estimated class retain probability for easy and difficult test instances respectively. The green \textsf{o}'s denote the ground truths, while the red \textsf{x}'s denote the base model predictoins. (c) shows approximate posterior $q(\bz|\bx,\by;\phi)$. (d) shows generated retain masks from (b). (e) shows the average retain probability per class for hard instances. (f) shows sampled predictive distributions of easy and difficult instance respectively. 
		}
		\vspace{-0.1in}
	\end{figure*}
	
	Figure~\ref{easy_examples} shows the retain probabilities estimated for easy examples, in which case the model set the retain probability to be high for the true class, and evenly low for non-target classes. Thus, when the examples are easy, the dropout probability estimator works like a second classifier. However, for difficult examples in Figure~\ref{p_examples} that is missclassified by the base softmax function, we observe that the retain probability is set high for the target class and few other candidates, as this helps the model focus on the classification between them. For example, in Figure \ref{p_examples}, the instance from class 3 sets high retain probability for class 5, since its handwritten character looks somewhat similar to number 5.
	However, the retain probability could be set differently even across the instances from the same class, which makes sense since even within the same class, different instances may get confused with different classes. For example, for the first instance of 4, the class with high retain probability is 0, which looks like 0 in its contour. However, for the second instance of 4, the network sets class 9 with high retain probability as this instance looks like 9.
	
	\begin{figure*}[t]
		\centering
		\includegraphics[height=3cm]{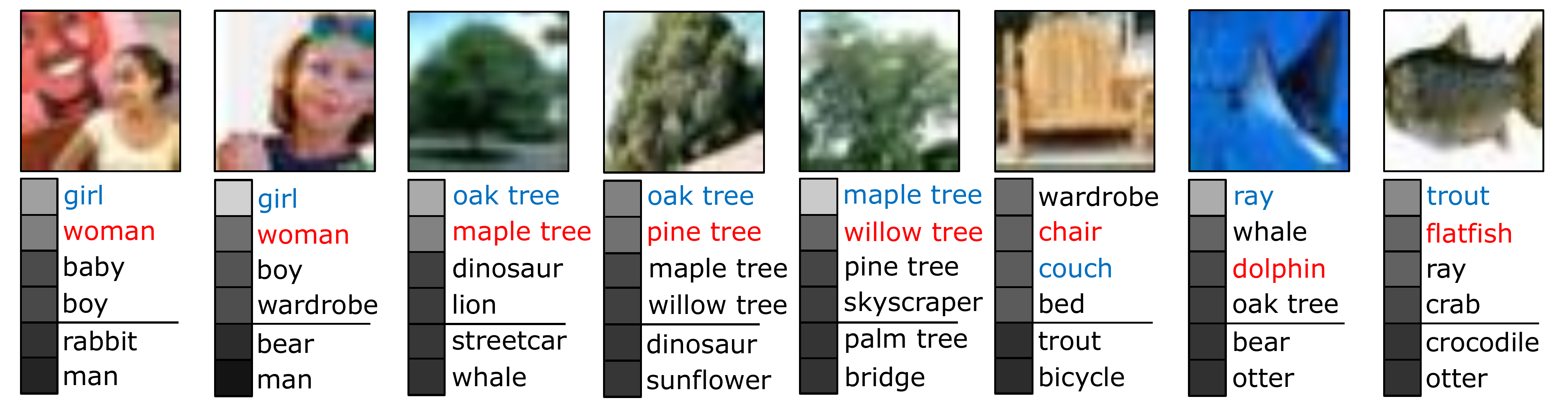}
		\vspace{-0.05in}
		\caption{\small Examples from CIFAR-100 dataset with top-$4$ and bottom-$2$ retain probabilities. Blue and red color denotes the ground truths and base model predictions respectively.}\label{c100_examples}
		\vspace{-0.11in}
	\end{figure*}

	Similar behaviors can be observed on CIFAR-100 dataset (Figure~\ref{c100_examples}) as well. As an example, for instances that belong to class \emph{girl}, DropMax sets the retain probability high on class \emph{woman} and \emph{boy}, which shows that it attends to most confusing classes to focus more on such difficult problems.
	
	We further examine the class-average dropout probabilities for each class in MNIST dataset in Figure~\ref{B}. We observe the patterns by which classes are easily confused with the others. For example, class 3 is often confused with 5, and class 4 with 9. It suggests that retain probability implicitly learns correlations between classes, since it is modeled as an input dependent distribution. Also, since DropMax is a Bayesian inference framework, we can easily obtain predictive uncertainty from MC sampling in Figure \ref{uncertainty}, even when probabilistic modeling on intermediate layers is difficult.

	\section{Conclusion and Future Work}
	\vspace{-0.05in}
	We proposed a stochastic version of a softmax function, DropMax, that randomly drops non-target classes at each iteration of the training step. DropMax enables to build an ensemble over exponentially many classifiers that provide different decision boundaries. We further proposed to learn the class dropout probabilities based on the input, such that it can consider the discrimination of each instance against more confusing classes. We cast this as a Bayesian learning problem and present how to optimize the parameters through variational inference, while proposing a novel regularizer to more exactly estimate the true posterior. We validate our model on multiple public datasets for classification, on which our model consistently obtains significant performance improvements over the base softmax classifier and its variants, achieving especially high accuracy on datasets for fine-grained classification. For future work, we plan to further investigate the source of generalization improvements with DropMax, besides increased stability of gradients (Appendix B).

	\section*{Acknowledgement}
	\vspace{-0.1in}
	This research was supported by the Engineering Research Center Program through the National Research Foundation of Korea (NRF) funded by the Korean Government MSIT (NRF-2018R1A5A1059921), Samsung Research Funding Center of
	Samsung Electronics (SRFC-IT150203), Machine Learning and Statistical Inference Framework for Explainable Artificial Intelligence (No.2017-0-01779), and Basic Science Research Program through the National Research Foundation of Korea (NRF) funded by the Ministry of Education (2015R1D1A1A01061019).
	Juho Lee's research leading to these results has received funding from the European Research Council under the European Union’s Seventh Framework Programme (FP7/2007-2013) ERC grant agreement no. 617071.
	
	\bibliographystyle{abbrv}
	\bibliography{refs}
	
	\appendix
	
	\section{Justification of the Observation 2.}
	Here we provide the justification and intuition of the observation 2 in Section 4.3 of the main paper. 
	\paragraph{Observation 2.}
	\textit{The true posterior of the target retain probability $p(z_t=1|\bx,\by)$ is $1$, if we exclude the case $z_1=z_2=\dots=z_K=0$, i.e. the retain probability for every class is $0$.}
	
	To verify it, we first need to understand what it means by saying $z_t=0$ even after the observation of the target $\by$. Firstly, suppose that the target mask $z_t=0$ and there exists at least one nontarget mask $z_{j\neq t}=1$. Then, the corresponding likelihood and the true posterior becomes
	\begin{align}
	p(\by|\bx,z_t=0, \bz_{\backslash t}) &= \frac{(0+\varepsilon)\exp(o_t)}{(1+\varepsilon)\exp(o_j) + \sum_{k \neq j}(z_k+\varepsilon)\exp(o_k)} \approx 0 \\
	p(z_t=0,\bz_{\backslash t}|\bx,\by) &= p(\by|\bx,z_t=0, \bz_{\backslash t})\frac{p(z_t=0, \bz_{\backslash t}|\bx)}{p(\by|\bx)} \approx 0 \label{eq:zero1}
	\end{align}
	where $\varepsilon > 0$ is a sufficiently small constant (e.g. $10^{-20}$). In other words, after knowing which class is the target, it is impossible to reason that the target class has been dropped out while some nontarget classes have not. 
	
	Secondly, suppose $z_t=0$ and $\bz_{\backslash t}=\mathbf{0}$. Then the likelihood and the true posterior becomes
	\begin{align}
	p(y_t=1|\bx,\bz=\mathbf{0}) &= \frac{(0+\varepsilon)\exp(o_t)}{\sum_k(0+\varepsilon)\exp(o_k)} = \frac{\exp(o_t)}{\sum_k\exp(o_k)} > 0 \\
	p(\bz=\mathbf{0}|\bx,\by) &= p(\by|\bx,\bz=\mathbf{0})\frac{p(\bz=\mathbf{0}|\bx)}{p(\by|\bx)} \geq 0
	\label{eq:zero2} 
	\end{align}
	In other words, after observing the label, it is one of the possible scenarios that all the target and nontarget classs have been dropped out at the same time. Combining \eqref{eq:zero1} and \eqref{eq:zero2}, we can conclude that $z_t=0$ only if $\bz_{\backslash t}=\mathbf{0}$, given $\by$. Ohterwise, $z_t=1$.
	
	Then, how can we express this relationship with the approximate posterior $q(\bz|\bx,\by) = \prod_k q(z_k|\bx,\by)$? It is impossible because we do not consider the correlations between $z_1,\dots,z_K$ under the mean-field approximation. In such a case, if we allow $q(z_t|\bx,\by) < 1$ somehow while having no means to force $z_t=0 \rightarrow \bz_{\backslash t}=\mathbf{0}$, then whenever $z_t$ is realized to be $0$, we always see the devation from the true posterior by the amount $q(\bz_{\backslash t}|\bx,\by)$ deviates from $\prod_{k \neq t}\Ber(z_k; 0)$. It also causes severe learning instability since reverting $z_t$ back to $1$ requires huge gradients. Considering that the case $\bz_{\backslash t}=\mathbf{0}$, one of the $2^{K-1}$ combinations, is insignificant, we ignore this case and let $q(z_t|\bx,\by)=\Ber(z_t;1)$. Except that case, the solution exactly matches the true marginal posterior $p(z_t|\bx,\by)$.
	
	\section{Stability of Gradients}
	The effect of DropMax regularization can be also explained in the context of the stability of stochastic gradient descent (SGD)~\cite{Hardt16icml, Bousquet02jmlr}, where a stable algorithm is preferred
	to achieve small generalization error.
	Suppose that the current model correctly classifies an example with small
	confidence. 
	DropMax regularization incurs a penalty to restrict the model from classifying an example too much perfectly (i.e. $o_t \gg \max_{k \in [K] \backslash \{t\}} o_k$).
	This automatically suggests that the magnitude of gradients of DropMax at this example is smaller than that of softmax, which helps to prevent from over-fitting and generalize better, as discussed in~\cite{Hardt16icml}.
	
	Denoting $\psi=\{\bW, \bb\}$, we consider the expected cross entropy as our loss function:
	\begin{align}
	\sum_{i=1}^{N}l(\bx_i, \by_i;\psi) = \sum_{i=1}^N\E_{\bz_i}\Big[-\log p\left(y_{i, t}=1 | \bh_i, \bz_i; \psi \right)\Big],
	\label{eq:loss}
	\end{align}
	where $\bh_i = \nn(\bx_i;\omega)$ is the last feature vector of an arbitrary neural network,
	$\bz_i$ and $p(y_{i,t}=1|\bh_i, \bz_i; \psi)$ are defined in Eq. (5) in the main paper.
	We consider an example that is correctly classified with small confidence;
	\begin{cond}
		Suppose that we are given a labeled example $\bx_i$ and $\by_i$.
		We assume that the retain probabilities denoted by $\{\rho_k\}_{k=1}^K$ follow the case: For a target class $t$, $\rho_t$ is greater than $\max_{k \in [K] \backslash \{t\}}\rho_k$. For a non-target class $k$, $\rho_k$ is equal to the one of any non-target classes.
		\label{cond:retain_prob}
	\end{cond}
	We further assume that Bernoulli parameter for $\bz_i$ is fixed, but different for each example.
	For simplicity, we denote $o_k(\bx_i;\psi)$ as $o_k$ when the context is clear.
	
	We then decompose the expected loss into the standard cross entropy with softmax and the regularization term introduced by DropMax; 
	\begin{align}
	\sum_{i=1}^{N}&\Big(\widehat{l}(\bx_i, \by_i ; \psi) + 
	\calM(\bx_i, \by_i, \bz_i)\Big),
	\label{eq:decomp}
	\end{align}
	where $\widehat{l}(\bx_i, \by_i ; \psi) = -\log \frac{\exp(o_t)}{\sum_{k=1}^K\exp(o_k)}$ that is the standard cross-entropy loss with softmax and $\calM(\bx_i, \by_i, \bz_i)=
	\E_{\bz_i}\left[\log \frac{\sum_{k=1}^K(z_k+\epsilon)\exp(o_k)}{(z_t+\epsilon)\sum_{k=1}^{K}\exp(o_k)}\right]$.
	We derive the upper bound on the regularization term 
	by Jensen's inequality and keep terms only related to $\psi$;
	\begin{align}
	\sum_{i=1}^{N}\left[\log\sum_{k=1}^{K}\left(\rho_k + \epsilon)\exp(o_k\right) - \log\sum_{k=1}^{K}\left(\exp(o_k)\right)\right]
	\end{align}
	
	We now compute the magnitude of gradient of DropMax to show if it is smaller than the one of softmax, which helps to stabilize the learning procedure. For ease of analysis, we consider the gradient for a target class\footnote{We can make the similar arguments for non-target classes.}:
	\begin{align}
	\frac{\partial \calM(\bx_i, \by_i, \bz_i)}{\partial \bw_t} &\leq \left(\frac{(\rho_t + \epsilon)\exp(o_t)}{\sum_k (\rho_k + \epsilon)\exp(o_k)} - \frac{\exp(o_t)}{\sum_k \exp(o_k)} \right)\frac{\partial o_t}{\partial \bw_t} \\
	\frac{\partial \widehat{l}(\bx_i, \by_i; \psi)}{\partial \bw_t}
	&= \left(\frac{\exp(o_t)}{\sum_k\exp(o_k)} - 1 \right)\frac{\partial o_t}{\partial \bw_t}.
	\end{align}
	According to Condition~\ref{cond:retain_prob}, it is easy to see that
	\begin{align}
	0 < \left(\frac{(\rho_t + \epsilon)\exp(o_t)}{\sum_k (\rho_k + \epsilon)\exp(o_k)} - \frac{\exp(o_t)}{\sum_k \exp(o_k)} \right),
	\label{eq:inequality}
	\end{align}
	which suggests that the gradient direction of regularizer
	is opposite to that of $\widehat{l}(\bx_i, \by_i; \psi)$.
	For an example that can be correctly classified 
	with small margin, DropMax regularization incurs a penalty to restrict the model from classifying an example too much perfectly (i.e. $o_t \gg \max_{k \in [K] \backslash \{t\}} o_k$). This means that DropMax is relatively more stable 
	than softmax in the notion of magnitude of gradient,
	which helps to prevent from over-fitting
	and generalize better.
	
	The convergence plot of MNIST-55K dataset (Figure 3(a) in the main paper) supports agrees with our argument that DropMax generalizes better by improving the stability of learning. Once the retain probabilies are trained to some degree and can roughly classify target and nontarget classes with minimum risk, then the burden to the softmax classifier is lessened, resulting in more stable gradients for the main softmax classifier.
	
	\section{Experimental Setup}
	Here we explain the experimental setup for the each dataset. 
	
	\textbf{1) MNIST.}\ The batchsize is set to $50$ and the training epoch is set to $2000$, $500$, and $100$ for $1K$,$5K$, and $55K$ dataset, respectively. We use Adam optimizer \cite{adam}, with learning rate starting from $10^{-4}$.
	The $\ell_2$ weight decay parameter is searched in the range of $\{0, 10^{-5}, 10^{-4}, 10^{-3}\}$. All the hyper-parameters are tuned with a holdout set. 
	
	\textbf{2) CIFAR-10.}\ We set batchsize to $128$ and the number of training epoch to $200$. We use stochastic gradient descent (SGD) optimizer with $0.9$ momentum. Learning rate starts from $0.1$ and multiplied by $0.1$ at $80, 120, 160$ epochs.
	The $\ell_2$ weight decay parameter is fixed at $10^{-4}$. 
	
	\textbf{3) CIFAR-100.}\ We used the same setup as CIFAR-10.
	
	\textbf{4) AWA.}\ Batchsize is set to $125$ and the number of training epochs is set to $300$. We use SGD optimizer with $0.9$ momentum. Learning rate starts from $10^{-2}$, and is multiplied by $0.1$ at $150$ and $250$ epochs. Weight decay is set to $10^{-4}$. 
	
	\textbf{5) CUB-200-2011.}\ Batchsize is set to $125$ and the number of training epochs is set to $400$. SGD optimizer with $0.9$ momentum is used. Learning rate starts from $10^{-2}$ and is multiplied by $0.1$ at $200$ and $300$ epochs. We set the weight decay to $10^{-3}$ which is bigger than the other datasets, considering that the size of the dataset is small compared to the network capacity.
	
	\begin{figure*}
		\centering
		\includegraphics[width=1.0\linewidth]{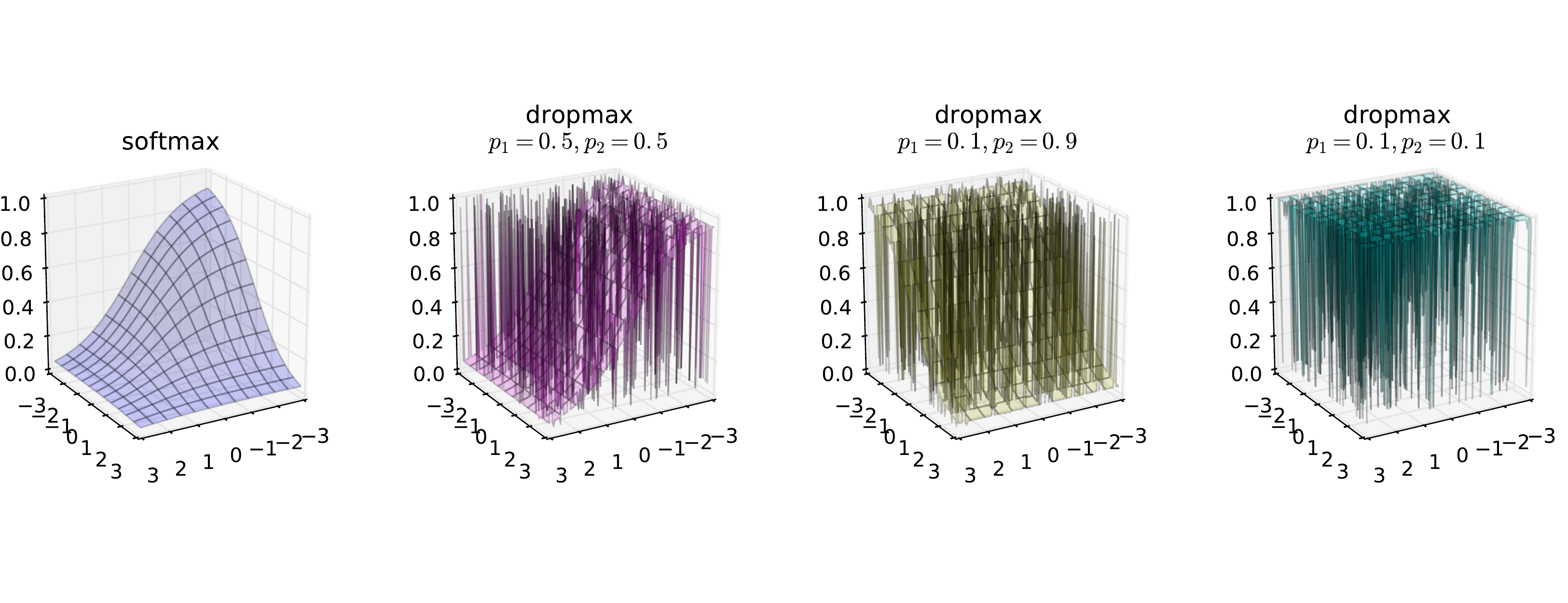}
		\vspace{-0.6in}
		\caption{Contour plots of softmax and DropMax with different retain probabilities. For DropMax, we sampled the Bernoulli variables
			for each data point with fixed probabilities.}\label{contour}
	\end{figure*}
	\begin{figure*}[t]
		\subfigure[]{\includegraphics[height=3.5cm]{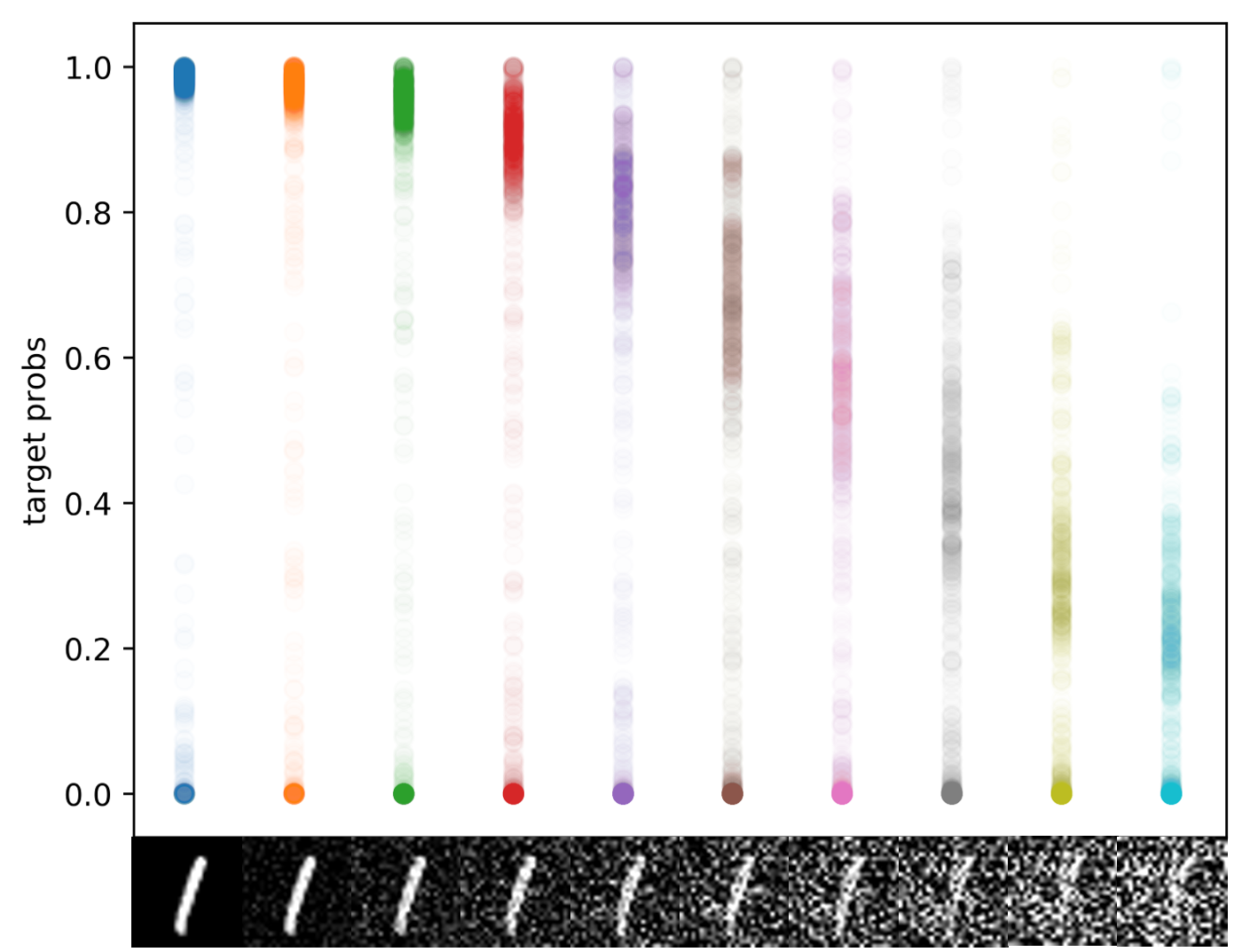}\label{uncertainty1}}
		\subfigure[]{\includegraphics[height=3.5cm]{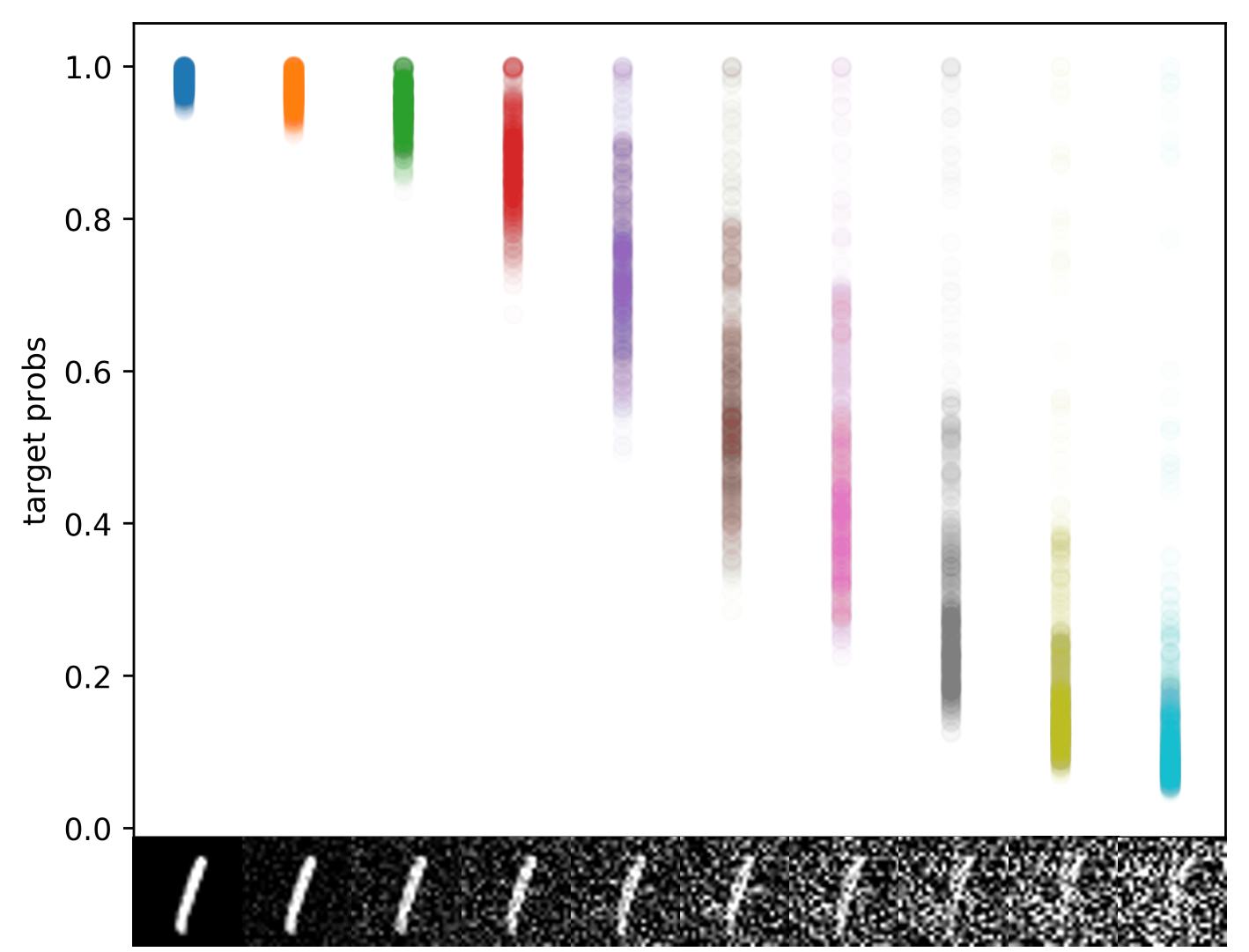}\label{uncertainty2}} 
		\subfigure[]{\includegraphics[height=3.5cm]{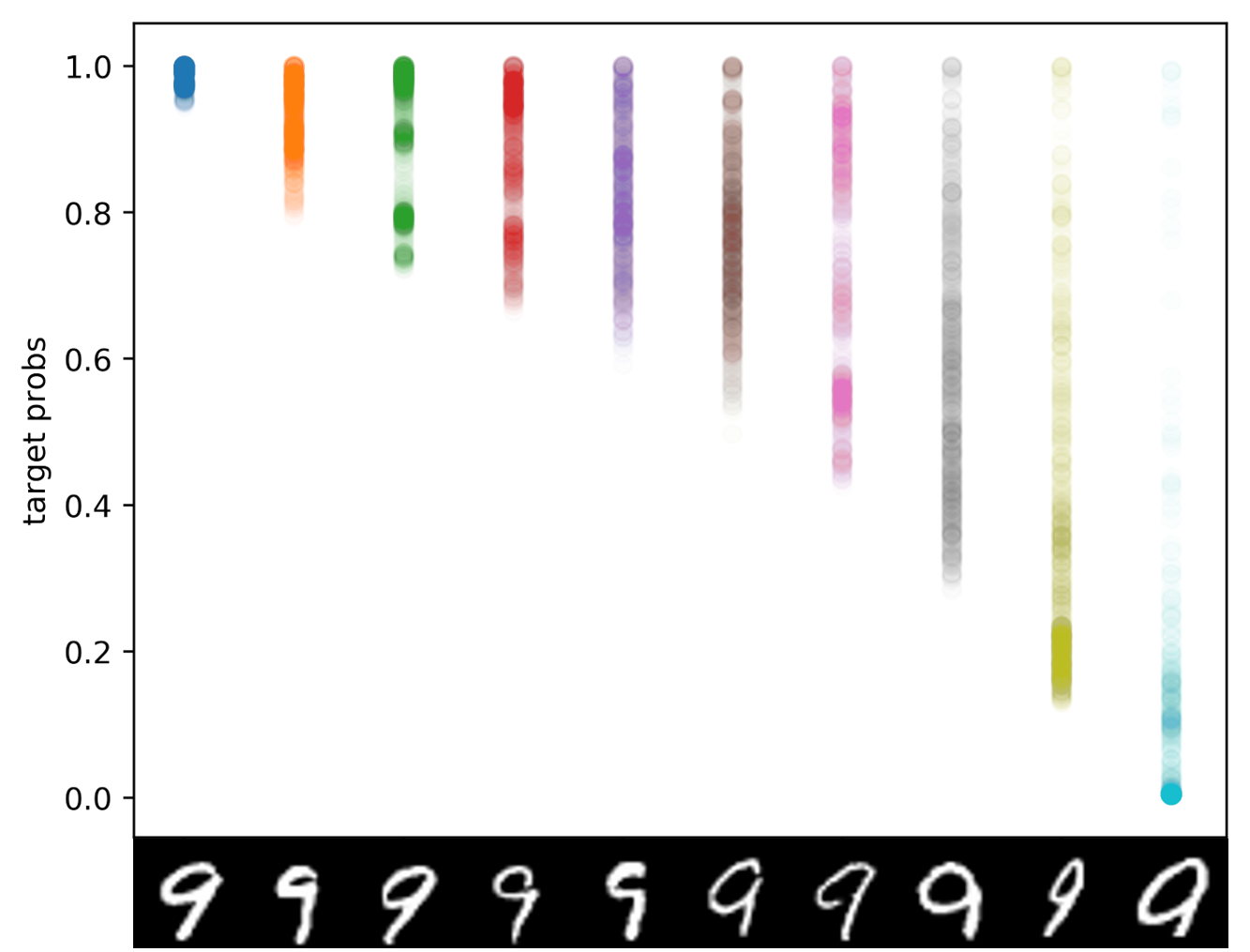}\label{uncertainty3}} 
		\vspace{-0.1in}
		\caption{\small (a) Monte-Carlo sampling of the target probabilities ($S=1000$) w.r.t. the different amount of noise on an instance from class 1. (b) Same as (a), except we do not sample the target mask to reduce the unnecessary variances (simply replace $z_t\sim \bern(\rho_t)$ with $\rho_t$). (c) MC sampling with real examples having different level of difficulties.}
		\vspace{-0.1in}
	\end{figure*}
	
\end{document}